\pdfoutput=1

\documentclass[11pt]{article}

\usepackage[review]{acl}

\usepackage{times}
\usepackage{latexsym}
\usepackage{times}
\usepackage{latexsym}
\usepackage{listings}
\usepackage{adjustbox}
\usepackage{amsmath}
\usepackage{breqn}
\usepackage{booktabs}
\usepackage{multirow}
\usepackage{multicol}
\usepackage{algorithmic}
 \usepackage{float}
\usepackage{hyperref}
\usepackage{tikz}
\usepackage{tcolorbox}
\usepackage{xcolor}
\usepackage[linesnumbered,lined,boxed,commentsnumbered]{algorithm2e}
\usepackage[T1]{fontenc}

\usepackage[utf8]{inputenc}

\usepackage{microtype}

\usepackage{inconsolata}

\title{Abstract Meaning Representation-Based Logic-Driven Data Augmentation for Logical Reasoning}

\author{Qiming Bao, Alex Yuxuan Peng, Zhenyun Deng, Wanjun Zhong, Gael Gendron, \\ \textbf{Timothy Pistotti, Neset Tan, Yang Chen, Yonghua Zhu,} \\ \textbf{Paul Denny,} \textbf{Michael Witbrock,} \and \textbf{Jiamou Liu} \\
        Address line \\ ... \\ Address line}

\author {
    Qiming Bao\textsuperscript{\rm 1, 2}, Alex Yuxuan Peng\textsuperscript{\rm 1}, Zhenyun Deng\textsuperscript{\rm 3}, Wanjun Zhong\textsuperscript{\rm 4}, \\ \textbf{Gaël Gendron}\textsuperscript{\rm 1}, \textbf{Timothy Pistotti}\textsuperscript{\rm 1}, \textbf{Neşet Tan}\textsuperscript{\rm 1}, \textbf{Nathan Young}\textsuperscript{\rm 1}, \textbf{Yang Chen}\textsuperscript{\rm 1}, \\ \textbf{Yonghua Zhu}\textsuperscript{\rm 1}, \textbf{Paul Denny}\textsuperscript{\rm 5}, \textbf{Michael Witbrock}\textsuperscript{\rm 1}, \and \textbf{Jiamou Liu}\textsuperscript{\rm 1}\\
    \textsuperscript{\rm 1}Strong AI Lab, NAOInstitute, Waipapa Taumata Rau - The University of Auckland\\
    \textsuperscript{\rm 2}Xtracta, New Zealand\\
    \textsuperscript{\rm 3}Department of Computer Science and Technology, University of Cambridge, UK\\
    \textsuperscript{\rm 4}School of Computer Science and Engineering, Sun Yat-Sen University, China\\
    \textsuperscript{\rm 5}School of Computer Science, The University of Auckland, New Zealand\\
    \{qbao775,ypen260,ntan607,yche767,ggen187\}@aucklanduni.ac.nz, zd302@cam.au.uk \\
}

\begin{document}
\nolinenumbers
\maketitle
\begin{abstract}
Combining large language models with logical reasoning enhances their capacity to address problems in a robust and reliable manner. Nevertheless, the intricate nature of logical reasoning poses challenges when gathering reliable data from the web to build comprehensive training datasets, subsequently affecting performance on downstream tasks. To address this, we introduce a novel logic-driven data augmentation approach, AMR-LDA. AMR-LDA converts the original text into an Abstract Meaning Representation (AMR) graph, a structured semantic representation that encapsulates the logical structure of the sentence, upon which operations are performed to generate logically modified AMR graphs. The modified AMR graphs are subsequently converted back into text to create augmented data. Notably, our methodology is architecture-agnostic and enhances both generative large language models, such as GPT-3.5 and GPT-4, through prompt augmentation, and discriminative large language models through contrastive learning with logic-driven data augmentation. Empirical evidence underscores the efficacy of our proposed method with improvement in performance across seven downstream tasks, such as reading comprehension requiring logical reasoning, textual entailment, and natural language inference. Furthermore, our method leads on the ReClor leaderboard\footnote{\url{https://eval.ai/web/challenges/challenge-page/503/leaderboard/1347}}. The source code and data are publicly available\footnote{\href{https://github.com/Strong-AI-Lab/Logical-Equivalence-driven-AMR-Data-Augmentation-for-Representation-Learning}{AMR-LDA GitHub Repository}}.
\end{abstract}

\section{Introduction}

Enabling pre-trained large language models (LLMs) to reliably perform logical reasoning 
is an important step towards strong artificial intelligence \cite{DBLP:journals/corr/abs-1911-01547}.
However, data annotation for logical reasoning tasks is a difficult, time-consuming and costly process that has led to the scarcity of large-scale logical reasoning datasets derived from natural language on the web. Therefore, LLMs are usually trained on generic corpora or smaller logical reasoning datasets that lead to poor generalisation~\cite{wang-etal-2022-logic}. 
Automatic augmentation of logical reasoning data has the potential to enhance the generalisation and performance of LLMs on logical reasoning tasks. 

\begin{figure}[ht]
\centering
\includegraphics[width=0.93384\columnwidth]{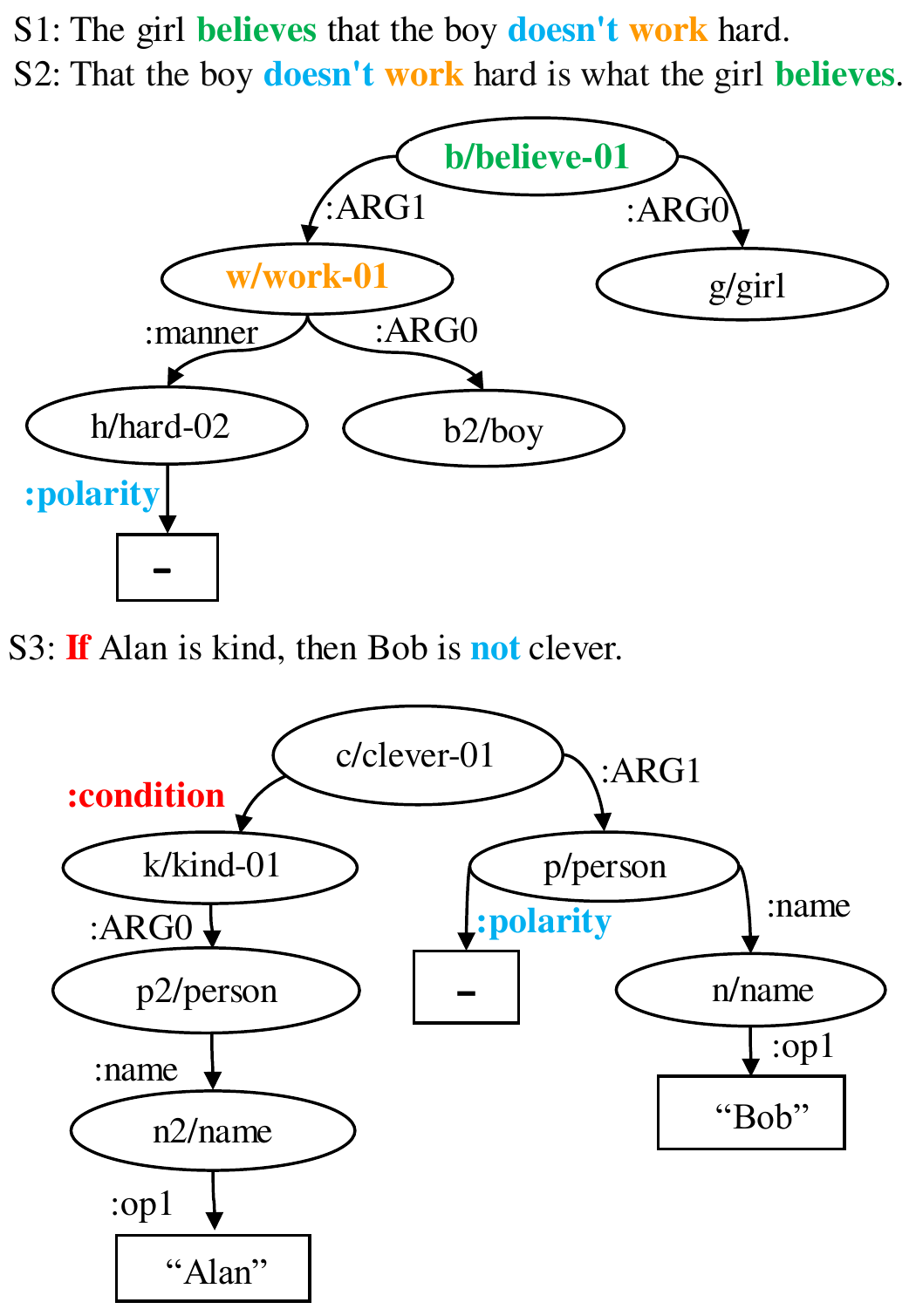}
\caption{
An example of AMR. Two sentences with the same semantic meaning can be represented as the same AMR graph. ``b'', ``g'', and ``w'' are variables. ``w/work-01'' refers to the variable ``w'' has an instance relation with the AMR concept ``work-01''. ``work'' is the frame from Propbank \cite{paul2002treebank} and ``-01'' is the sense of frame. ``:ARG0'', ``:ARG1'', ``:condition'', ``:polarity'' are frame arguments, following PropBank instructions. ``:condition'' and ``:polarity -'' are used to represent conditional and negative relationships.}
\label{amr-example}
\end{figure}

To address this challenge, we propose a logic-driven data augmentation method based on Abstract Meaning Representation (AMR). AMR is a structural representation of the semantics and logical structure of text via a rooted directed acyclic graph (DAG) \cite{shou2022amr}. Figure~\ref{amr-example} shows an example of an AMR graph. The AMR graph can be easily modified by changing nodes or arguments to create logically equivalent or nonequivalent graphs. 
By taking advantage of the ease of logical manipulation of AMR graphs and of end-to-end conversion between natural language and AMR graphs, our proposed data augmentation is not task-specific or template-dependent, and can generate logically equivalent and nonequivalent sentences that are diverse in their use of language.

In order to improve the performance of LLMs on downstream tasks requiring logical reasoning, we investigate two different applications of the proposed logic-driven data augmentation for two different types of language models. In this paper, we describe models such as RoBERTa \cite{liu2019roberta} and DeBERTa \cite{he2020deberta} as discriminative large language models, and models like GPT-3.5 \cite{openai2023chatgpt} as generative LLMs. 
We improve the reasoning ability of discriminative large language models by applying contrastive learning to identify logically equivalent and nonequivalent sentence pairs generated using the proposed data augmentation before fine-tuning the model further on downstream tasks. In order to improve the performance of generative LLMs on logical reasoning tasks without fine-tuning, we augment the input prompt by extending the question context and options using data augmentation. We summarize the paper's key contributions as follows:

\begin{enumerate}
    \item We propose an AMR-based logic-driven data augmentation method to automatically construct logically equivalent/nonequivalent sentences.
    \item We enhance the logical reasoning of large language models through logical-equivalence-identification contrastive learning and prompt augmentation.
    \item The experimental results show that our method can improve large language models' performance on downstream tasks including logical reasoning, textual entailment and natural language inference. 
\end{enumerate}

\section{Related Work}

Logical reasoning is rigorous thinking to derive a conclusion based on a given premise~\cite{seel2011encyclopedia,bronkhorst2020logical}. Existing reasoning datasets' reasoning can be categorised into two levels: sentence level, including tasks like natural language inference that assess if one sentence logically follows from another (e.g., MNLI~\cite{williams2017broad}, RTE~\cite{wang2018glue}, MRPC~\cite{dolan2005automatically}, QNLI~\cite{rajpurkar2016squad}, QQP~\cite{wang2018glue}); passage level, which requires logical deduction from given contexts, questions, and multiple choices (e.g., PARARULE~\cite{clark2020transformers}, PARARULE-Plus~\cite{bao2022multi}) and reading comprehension tasks (e.g., ReClor~\cite{yu2020reclor}, LogiQA~\cite{liu2020logiqa}). We introduce an abstract meaning representation-based methodology for logic-driven data augmentation aimed at enhancing models' logical reasoning capabilities across these tasks.

There are three primary methods for enhancing the capabilities of pre-trained language models in logical reasoning and general natural language understanding: 1) Data augmentation with fine-tuning, exemplified by AMR-DA~\cite{shou2022amr}, which employs Abstract Meaning Representation for paraphrasing, and LReasoner~\cite{wang-etal-2022-logic}, which uses templates and syntax parsing for constructing logically equivalent sentences; 2) Continual pre-training, with methods like MERIt~\cite{jiao2022merit} integrates a meta-path strategy for discerning logical text structures and a counterfactual data augmentation strategy to preclude pre-training shortcuts. IDoL~\cite{xu-etal-2023-idol} utilises six logical indicators~\cite{pi2022logigan,prasad2008penn} to build a logic pre-training dataset from Wikipedia, enhancing the logical reasoning capabilities of pre-trained models. 3) Prompting, notably Chain-of-Thought prompting~\cite{wei2022chain}, to improve multi-step logical reasoning performance. Our AMR-LDA surpasses LReasoner-LDA by incorporating a broader range of logical equivalence laws, enabling the automatic construction of more precise logically equivalent sentences. Our contrastive learning method enhance the performance of pre-trained models, including MERIt and IDoL, on logical reasoning tasks. Additionally, our AMR-based logic-driven prompt augmentation can improve large language models' logical reasoning capabilities, contrasting with the detrimental effects of CoT Prompting and AMR-DA.

\section{Method}\label{sec:method}
\subsection{System Architecture}
\begin{figure*}[ht]
\centering
\includegraphics[width=\textwidth]{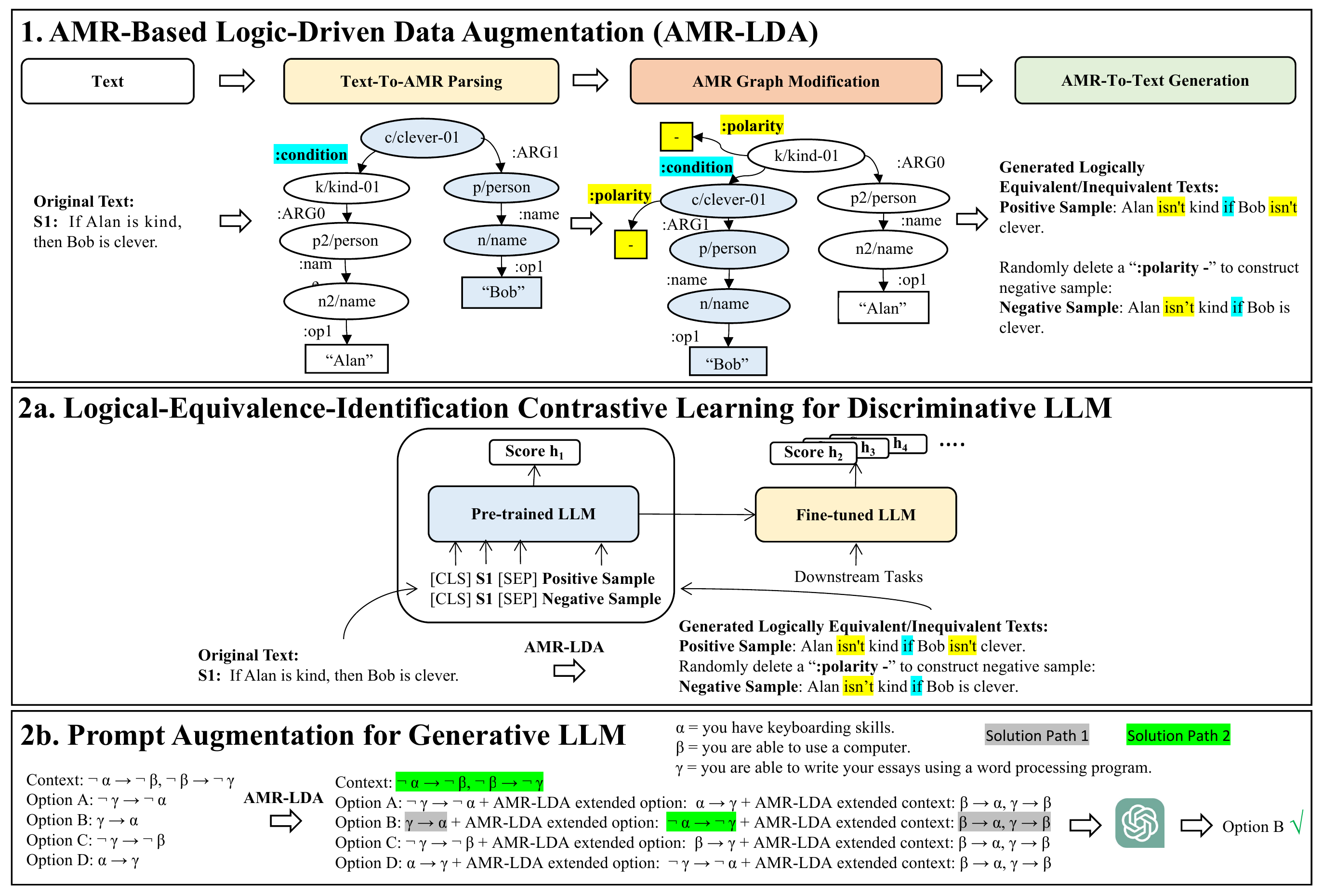}
\caption{Architecture of AMR-LDA (1) and its applications to improve the reasoning performance of discriminative LLMs with contrastive learning (2a) and autoregressive generative LLMs by augmenting input prompts without fine-tuning (2b).}
\label{system-architecture}
\end{figure*}

Our system, shown in Figure \ref{system-architecture}, features an \textbf{AMR-Based Logic-Driven Data Augmentation Module} that parses sentences into AMR graphs, modifies the graphs to generate corresponding logically equivalent and nonequivalent graphs, then converts these back into natural language. The \textbf{Logical-Equivalence-Identification Contrastive Learning Module} aims to improve the logical reasoning ability of discriminative large language models by conducting contrastive learning to identify equivalent and nonequivalent sentence pairs, before further fine-tuning the model on downstream tasks. The \textbf{Prompt Augmentation Module} is intended to improve the performance of generative autoregressive LLMs on logical reasoning tasks by applying the data augmentation module to the input fed into the models at inference time, without performing any fine-tuning.





\subsection{AMR-Based Logic-Driven Data Augmentation}\label{sec:amr-lda}
We propose \textbf{A}bstract \textbf{M}eaning \textbf{R}epresentation-based \textbf{L}ogic-driven \textbf{D}ata \textbf{A}ugmentation (\textbf{AMR-LDA}) to construct logically equivalent and nonequivalent sentences automatically. For simplicity, we consider only individual sentences, and propositional logic statements expressed in natual language. AMR-LDA involves the following steps: \textit{\textbf{1)}}: Convert a sentence into AMR graph. \textit{\textbf{2)}}: Logically augment the AMR graph. \textit{\textbf{3)}}: Convert the logically augmented AMR graph back into natural language. 

\begin{table*}[h]
\centering
\renewcommand{\arraystretch}{1}
\resizebox{0.8\linewidth}{!}{
\begin{tabular}{@{}lcc@{}}
\toprule
Original sentence                                                                                & Positive sample                                                                             & Negative sample                                                                                     \\ \midrule
\multirow{2}{*}{\begin{tabular}[c]{@{}l@{}}If Alan is kind, \\ then Bob is clever.\end{tabular}} & Alan \colorbox{yellow}{\color{black}isn't} kind \colorbox{cyan}{\color{black}if} Bob \colorbox{yellow}{\color{black}isn't} clever.                                                        & Alan \colorbox{yellow}{\color{black}isn't} kind \colorbox{cyan}{\color{black}if} Bob \colorbox{yellow}{\color{black}is} clever.                                                                   \\ \cmidrule(l){2-3} 
                                                                                                 & Alan is \colorbox{yellow}{\color{black}not} kind \colorbox{yellow}{\color{black}or} Bob is clever.                                                          & Alan is kind \colorbox{yellow}{\color{black}or} Bob is clever.                                                                      \\ \midrule
The bald eagle is strong.                                                                        & The bald eagle is \colorbox{yellow}{\color{black}not weak}.                                                                 & The bald eagle is \colorbox{yellow}{\color{black}weak}.                                                                             \\ \midrule
\begin{tabular}[c]{@{}l@{}}The bald eagle is clever \\ and the wolf is fierce.\end{tabular}      & \begin{tabular}[c]{@{}l@{}}\colorbox{cyan}{\color{black}The wolf is fierce} and \\ \colorbox{cyan}{\color{black}the bald eagle is clever}.\end{tabular} & \begin{tabular}[c]{@{}l@{}}\colorbox{cyan}{\color{black}The wolf is} \colorbox{yellow}{\color{black}not} \colorbox{cyan}{\color{black}fierce} and \\ \colorbox{cyan}{\color{black}the bald eagle is} \colorbox{yellow}{\color{black}not} \colorbox{cyan}{\color{black}clever}.\end{tabular} \\ \bottomrule
\end{tabular}}
\caption{Examples of generated logically equivalent (positive) and nonequivalent sentences(negative). The blue background highlights the parts of the original sentence that have been moved from their original positions. The yellow background highlights the change in polarity from the original sentence.}
\label{first-stage-finetuning}
\end{table*}

\paragraph{Text-To-AMR Parsing}
A text-to-AMR model is used to parse a sentence into an AMR graph. In this step, the input is a natural language sentence written in English. The output is a rooted, labeled, directed, and acyclic AMR graph that captures the main semantic information of the sentence.

\paragraph{AMR Graph Modification}
The AMR graph is modified to construct logically equivalent and nonequivalent graphs. To create logically equivalent graphs, we consider four different logical equivalence laws: \textit{double negation}, \textit{commutative}, \textit{implication}, and \textit{contraposition} laws. These laws of logical equivalence are defined below using propositional statements $\mathcal{A}$ and $\mathcal{B}$, followed by examples in natural language (e.g. $\mathcal{A}$ is  ``Alan is kind'' and $\mathcal{B}$ is ``Bob is clever'').
\bigskip   

\paragraph{Logical Equivalence}
Logical equivalence is a fundamental concept in formal logic~\cite{mendelson2009introduction}. It can be formally defined as: Two propositions or statement forms $P$ and $Q$ are logically equivalent if they have the same truth value in every possible circumstance, or in every possible model. This can be denoted as $P \equiv Q$. This condition can also be described by the statement: $P$ and $Q$ are logically equivalent if and only if the statement ``P if and only if Q'' is a tautology. A tautology is a statement that is always true, regardless of the truth values of its components. In terms of truth tables, $P$ and $Q$ are logically equivalent if their truth tables are identical, i.e., $P$ and $Q$ have the same truth value for each possible assignment of truth values to their components. 

\noindent\textbf{Definition 1: Contraposition Law} 
$$
(\mathcal{A} \rightarrow \mathcal{B})  \Leftrightarrow (\neg \mathcal{B} \rightarrow \neg \mathcal{A})
$$
{\em If Alan is kind, then Bob is clever. $\quad \Leftrightarrow \quad$ If Bob is not clever, then Alan is not kind.}
\bigskip  

To implement the contraposition law, we first swap the first half of the sentence with the second half if the AMR parser detects that the sentence is a conditional statement (e.g. ``if-then'', as marked by the blue background in Table \ref{first-stage-finetuning}). In the second step, we construct logically equivalent sentences for the four potential scenarios in which the negation may appear. Here, we use one such scenario as an example. If the first half of the sentence has no negation and the second half of the sentence has no negation either, then we will add the negative polarity argument, ``:polarity -'', to the first half and the second half of the sentence to construct logically equivalent sentences (marked with the yellow background in Table \ref{first-stage-finetuning}). AMR uses ``:polarity -'' to represent negation (e.g. ``not''). Note that our method is not limited to the word ``not'', the negative argument ``: polarity -'' in the AMR graph may represent other negative words in the original sentence. We discuss those cases in Section~\ref{eq:double_negation} Definition 4 when describing the implementation for double negation law. An example of the augmentation process be found in Figure~\ref{double-negation-example} in Appendices. 
\bigskip  

\noindent\textbf{Definition 2: Implication Law} 
$$
(\mathcal{A} \rightarrow \mathcal{B}) \Leftrightarrow(\neg \mathcal{A} \vee \mathcal{B})
$$
{\em If Alan is kind, then Bob is clever. $\Leftrightarrow$ Alan is not kind or Bob is clever.}
\bigskip 

We consider two scenarios. If the sentence is detected by the AMR parser as a conditional statement, then we replace the conditional connective with a disjunction connective (marked with yellow background in Table~\ref{first-stage-finetuning}). In the second scenario, if the sentence contains disjunction connectives, we replace the disjunction connective with conditional connective and remove the negative polarity from the AMR graph if it exits. Otherwise, a negative polarity argument will be added. An example can be found in~\nameref{sec:appendix} Figure~\ref{contraposition-implication-example}.
\bigskip  

\noindent\textbf{Definition 3: Commutative Law}
$$
(\mathcal{A} \wedge \mathcal{B})  \Leftrightarrow(\mathcal{B} \wedge \mathcal{A})
$$
{\em Alan is kind and Bob is clever. $\quad \Leftrightarrow \quad$ Bob is clever and Alan is kind.}
\bigskip  

If the AMR graph has a conjunction connective, we swap the order of the first half of the graph with the second half. An example can be found in Table~\ref{first-stage-finetuning} and in \nameref{sec:appendix} Figure \ref{commutative-example}. The sub-sentence ``The wolf is fierce'' and ``the bald eagle is clever'' marked as blue have been swapped.
\bigskip  

\noindent\textbf{Definition 4: Double Negation Law} 
\begin{equation*}\label{eq:double_negation}
\mathcal{A} \Leftrightarrow \neg \neg \mathcal{A}
\end{equation*}
{\em It is raining. $\Leftrightarrow$ It is not the case that it is not raining.}
\bigskip 

We apply the double negation law only to those sentences and their AMR graphs that do not contain the ``:polarity -'' argument which represents negative polarity. There are several words that can be represented as ``:polarity -'', such as ``not'', ``no'', ``never'', ``none'', and ``nothing''. A representative example can be seen in Table~\ref{first-stage-finetuning} and in \nameref{sec:appendix} Figure \ref{double-negation-example}. The original sentence is ``The bald eagle is strong''. The logically equivalent sentence we construct using the double negation law is ``The bald eagle is not weak'', while the logically nonequivalent sentence is ``The bald eagle is weak''. Note that the generated sentences do not contain the word ``not'' twice. We avoid generating sentences with ``not'' appearing multiple times consecutively because they are uncommon and unnatural. The process of applying double negation law is as follows: convert the sentence into an AMR graph; augment the AMR graph by adding a negative polarity argument ``: polarity -''; convert the modified AMR graph back into a natural language sentence; lastly, replace the adjective word with its antonym by using WordNet \cite{miller1995wordnet}.
To create logically nonequivalent sentences, we randomly delete or add a negative polarity argument ``:polarity -'' in the AMR graph. Additionally, we randomly sample a sentence from the corpus and consider it as logically nonequivalent to the original sentence. 

\paragraph{AMR-To-Text Generation}
Lastly, an AMR-to-text model is used to convert the modified AMR graph back into natural language, to generate a sentence that is logically equivalent or nonequivalent to the original sentence.

\subsection{Logical-Equivalence-Identification Contrastive Learning} \label{sec:contrastive-learning}
Inspired by SimCSE \cite{gao2021simcse} and SimCLR \cite{chen2020simple}, we propose to improve dicriminative language models' logical reasoning ability by performing contrastive learning to identify logically equivalent and nonequivalent sentence pairs that are generated using AMR-LDA (Figure~\ref{system-architecture}, 2a).

\paragraph{Contrastive Learning} The goal of contrastive learning is to minimise the distance of the hidden representations of two similar inputs while maximising the distance between two representations of dissimilar inputs. Our goal is to optimise the model to map logically equivalent sentences to hidden representations that are close to each other.
\begin{equation}
h\left(s, s^{+}\right) \gg h\left(s, s^{-}\right).
\end{equation}
$h$ is a score function used to measure the distance between two representations. $s$ is an original sentence, $s^{+}$ is a positive sample logically equivalent to the original sentence $s$, $s^{-}$ is a negative sample logically nonequivalent to the original sentence $s$. The expected semantic representation distance between $s$ and $s^{+}$ should be much closer than that of $s$ and $s^{-}$. The training loss can be written with the following formula:
\begin{equation}
    \mathcal{L}=-\sum \log \frac{\exp \left(h\left({+}\right)\right)}{\exp\left( h\left({+}\right)\right)+\exp \left(h\left({-}\right)\right)},
\end{equation}
where $h\left({+}\right)$ and $h\left({-}\right)$ are short for $h\left(s, s^{+}\right)$ and $h\left(s, s^{-}\right)$.

After the contrastive learning step, we further fine-tune the model on downstream tasks, including logical reasoning reading comprehension, natural language inference, and textual entailment.


\subsection{Prompt Augmentation}
To improve the performance of generative LLMs (e.g., GPT-3.5 or GPT-4) on logical reasoning tasks, we propose augmenting the input prompt using AMR-LDA before feeding it to the model (Figure~\ref{system-architecture}, 2b). In the example from Figure~\ref{system-architecture}, 2b, the context and options are marked in green and grey, respectively. The original Option B is ``If you are able to write your essays using a word processing program, then you have keyboarding skills,'' which cannot be explicitly inferred from the context without using the logical equivalence law (contraposition law). AMR-LDA is able to augment the original option and generate ``If you have no keyboarding skills, then you are not able to write your essays using a word processing program,'' which is logically equivalent to the original Option B, now also marked in green. This augmented Option B can be inferred from the given context. Furthermore, AMR-LDA is also applied to augmenting sentences within the context. The augmented, logically equivalent sentences from the context are ``If you are able to use a computer, then you have keyboarding skills. If you are able to write your essays using a word processing program, then you are able to use a computer,'' which are marked in grey and support the validity of the original Option B. Finally, the augmented option and context are combined and fed as a prompt to GPT-3.5/4. Based on the extended information, we can find two solution paths marked with grey and green backgrounds under \textbf{Module 2b} in Figure~\ref{system-architecture}. \textbf{\textit{Solution Path 1}} uses the sentence from the extended context marked with a grey background to support that Option B is correct. \textbf{\textit{Solution Path 2}} uses the sentence from the original context marked with a green background to support that the extended Option B is correct. Consequently, our method provides more solution paths for large language models to more effectively solve complex logical reasoning questions.


\section{Experiments}


\begin{table*}[h]
\centering
\renewcommand{\arraystretch}{1}
\setlength{\tabcolsep}{6pt}
\resizebox{1\linewidth}{!}{
\begin{tabular}{@{}lccccccccccc@{}}
\toprule
\multirow{2}{*}{Models/ Datasets} & \multicolumn{4}{c}{ReClor}                                            & \multicolumn{2}{c}{LogiQA}        & MNLI            & MRPC            & RTE             & QNLI            & QQP             \\ \cmidrule(l){2-12} 
                                  & Dev             & Test            & Test-E          & Test-H          & Dev             & Test            & \multicolumn{5}{c}{Eval}                                                                 \\ \midrule
RoBERTa                           & 59.73          & 53.20          & 72.57          & 37.97          & 35.43          & 34.50          & 88.95          & 90.44          & 83.39          & \textbf{94.73} & 90.89          \\
RoBERTa LReasoner-LDA             & 59.46          & 53.66          & 72.19          & 39.10          & 34.81          & 34.81          & 89.41          & 89.46          & 86.28          & 94.25          & 90.01          \\
RoBERTa AMR-DA                    & 58.66          & 53.93          & 66.81          & \textbf{43.80} & 36.45          & 37.22          & 89.74          & 90.44          & 86.28          & 94.42          & 92.06          \\
RoBERTa AMR-LDA                   & \textbf{65.26} & \textbf{56.86} & \textbf{77.34} & 40.77          & \textbf{40.29} & \textbf{38.14} & \textbf{89.78} & \textbf{90.93} & \textbf{86.64} & 94.49          & \textbf{93.14} \\\midrule
DeBERTaV2                         & 73.93          & 70.46          & 80.82          & 62.31          & 39.72          & 39.62          & 89.45          & 89.71          & 84.48          & 95.00          & \textbf{92.54} \\
DeBERTaV2 LReasoner-LDA           & 75.73          & 70.70          & 84.08          & 60.17          & 30.87          & 28.51          & 89.23          & 89.95          & 87.00          & 95.15          & 92.50          \\
DeBERTaV2 AMR-DA                  & 79.06          & 75.90          & 84.62          & 69.04          & 29.95          & 30.10          & \textbf{89.92} & 89.71          & 83.39          & 95.02          & 92.42          \\ 
DeBERTaV2 AMR-LDA                 & \textbf{79.40} & \textbf{77.63} & \textbf{85.75} & \textbf{71.24} & \textbf{42.34} & \textbf{39.88} & 89.67          & \textbf{90.20} & \textbf{88.09} & \textbf{95.24} & 92.47          \\\bottomrule
\end{tabular}}
\caption{Comparison between our proposed AMR-LDA and baseline models. We use RoBERTa-Large, DeBERTaV2-XXLarge as the pre-trained models. Our fine-tuned LLMs perform equally well or better than baseline methods.}
\label{main-synthetic-test}
\end{table*}

\subsection{Datasets}
\label{sec:datasets}
ReClor~\cite{yu2020reclor} and LogiQA~\cite{liu2020logiqa} are two challenging logical reasoning datasets. ReClor is collected from the Graduate Management Admission Test (GMAT) and the Law School Admission Test (LSAT). LogiQA is collected from the National Civil Service Examination~\cite{liu2020logiqa}. Additionally, we performed evaluations on five datasets for natural language inference and textual entailment tasks: MNLI~\cite{williams2017broad}, RTE~\cite{wang2018glue}, MRPC~\cite{dolan2005automatically}, QNLI~\cite{rajpurkar2016squad}, and QQP~\cite{wang2018glue}. MNLI, RTE, and MRPC assess the relationship between two sentences, while QNLI focuses on the relationship between a question and a sentence, and QQP evaluates the relationship between two questions.

\paragraph{Synthetic Data for Contrastive Learning}
In this paper, we performed contrastive learning for discriminative large language models on sentences augmented from a synthetic dataset. This dataset contains 14,962 sentences with different combinations of 23 entities, 2 relations and 40 attributes. Synthetic data was used to generate more controllable logical sentences. More details about the synthetic dataset can be found in the~\nameref{sec:appendix} Section~\ref{synthetic-data-construction}.

\subsection{Settings}
All experiments were conducted on 8 NVIDIA A100 GPUs, each with 80G of VRAM. Primary experiments on the ReClor and LogiQA datasets used three different random seeds; the average values are reported in Table~\ref{main-synthetic-test}. The parse\_xfm\_bart\_large and T5Wtense models from AMRLib\footnote{\url{https://amrlib.readthedocs.io/en/latest/models/}} were used for text-to-AMR and AMR-to-text conversions when generating logically augmented sentence pairs. The reason for selecting those two models is explained in subsection~\ref{preliminary_study}. In our experiments, RoBERTa~\cite{liu2019roberta} and DeBERTa~\cite{he2020deberta} were used as the discriminative large language models. We also compared our method against MERIt~\cite{jiao2022merit} and IDoL~\cite{xu-etal-2023-idol}, the leading models on the ReClor leaderboard. As for generative large language models, we applied GPT-3.5 (gpt-3.5-turbo)~\cite{openai2023chatgpt} and GPT-4~\cite{openai2023gpt4}. More details about the experiments, case studies and confidence intervals can be found in~\nameref{sec:appendix} Section~\ref{experiment_setup},~\ref{case-studies},~\ref{long-sentence-example}, and~\ref{confidence-intervals}.


\subsection{Logical-Equivalence-Identification Contrastive Learning for Discriminative LLMs} 
This section evaluates the effectiveness of contrastive learning on synthetic data augmented using AMR-LDA in order to improve the performance of discriminative language models on downstream tasks that require logical reasoning. We compare AMR-LDA against two baseline augmentation methods: AMR-DA~\cite{shou2022amr} and LReasoner-LDA~\cite{wang-etal-2022-logic}.
It is important to note that we do not use the whole system or pipeline from LReasoner, we only use the data augmentation method from LReasoner in our experiment. For each augmentation method, 14,962 pairs of logically equivalent and logically nonequivalent sentences are constructed with a positive to negative sample ratio of 1:1. Twenty percent of the augmented data are used as the validation set during contrastive learning. All the models are further fine-tuned and compared on downstream tasks requiring logical reasoning and natural language inference. The results as shown in Table~\ref{main-synthetic-test}, suggest that the models trained using AMR-LDA perform better in most cases compared with the other augmentation methods.

\subsection{Prompt Augmentation for Generative LLM} 
We adopt GPT-3.5 (gpt-3.5-turbo)~\cite{openai2023chatgpt} and GPT-4~\cite{openai2023gpt4} as the generative large language models for evaluating the effectiveness of prompt augmentation using AMR-LDA. The experiments are performed on the ReClor and LogiQA datasets. 
The experimental results are shown in Table~\ref{comparison-reclor-logiqa-chatgpt}. The models with prompt augmentation achieved better performance in all cases except for the ``hard'' test set for ReClor. We also compare our method against Chain-of-Thought Prompting (CoT)~\cite{wei2022chain} and AMR-DA~\cite{shou2022amr} for prompt augmentation. We apply AMR-DA to paraphrase each option and each sentence in the context, and the rest is the same as the AMR-LDA prompt augmentation. We found that CoT and augmentation with AMR-DA caused a decline in performance for both GPT-3.5 and GPT-4 in most cases, except for GPT-4 on LogiQA. The performance drop associated with using CoT Prompting has been reported by~\cite{xu-etal-2023-idol}. However, they only sampled 100 cases from the validation set, whereas we use the entire validation set and test set. AMR-DA conducts data augmentation by converting the text into an AMR graph and then randomly selecting one of the following operations: removing, swapping, substituting, or inserting an argument into the graph. The modified AMR graph is then converted back into a new sentence. This modification of the AMR may disrupt the original sentence’s structure and introduce noise into the prompt, potentially worsening performance. 

GPT-3.5 AMR-LDA performs better than GPT-3.5 on the general test set, which includes both test-E and test-H. The ReClor test set is hidden, so we do not have access to the detailed results for test-E and test-H. Therefore, we cannot provide a clear explanation as to why AMR-LDA seems to decrease the test-H metric for GPT-3.5. However, a detailed examination of the results reveals that GPT-3.5 achieves only a 0.5375 test accuracy on test-H, whereas GPT-4 attains a 0.8857 test accuracy on the same test. Furthermore, GPT-4 with AMR-LDA performs better on all the ReClor and LogiQA test sets. This suggests that GPT-3.5 might not be as effective in comprehending complex logical reasoning as GPT-4 and GPT-3.5 may understand augmented prompts poorly.


\begin{table}[h]
\centering
\renewcommand{\arraystretch}{1}
\resizebox{\columnwidth}{!}{
\begin{tabular}{@{}lcccccc@{}}
\toprule
\multirow{2}{*}{Models/Datasets} & \multicolumn{4}{c}{ReClor}                                                                             & \multicolumn{2}{c}{LogiQA}        \\ \cmidrule(l){2-7} 
                                 & Dev             & Test                       & Test-E                     & Test-H                     & Dev             & Test            \\ \midrule
GPT-3.5                      &   57.02        & 56.20 & 59.31 & \textbf{53.75} & 37.63          &   37.32        \\
+ CoT     &   34.80      &  25.80         &      27.50               &    24.46        &  23.96 & 24.57 \\
+ AMR-DA                      & 33.20        &  32.90         &     34.31                &    31.78        &  \textbf{40.55} & 31.49 \\
+ AMR-LDA                      &  \textbf{58.62}         & \textbf{56.69}            &        \textbf{60.90}              &    53.39         &  \textbf{40.55} & \textbf{39.47} \\
GPT-4                      & 87.35 &    89.60                 &     90.90                 &     88.57                 &   43.24        & 53.88         \\
+ CoT     &   37.00      &  24.80         &     26.13                &    23.75        & 23.50  & 27.03 \\
+ AMR-DA                      &   85.00      &  85.60         &       86.36              &    85.00        & 51.30  & 56.06 \\
+ AMR-LDA                        &  \textbf{87.73}          &    \textbf{90.20}                 &    \textbf{91.59}         &      \textbf{89.11}                &       \textbf{51.92}          &  \textbf{58.06}
               \\\bottomrule
\end{tabular}}
\caption{Comparison of Chain-of-Thought Prompting (CoT), AMR-DA, and AMR-LDA on GPT-3.5 and GPT-4, and between GPT-3.5 and GPT-4 alone, for evaluation on the ReClor and LogiQA test sets.}
\label{comparison-reclor-logiqa-chatgpt}
\end{table}

\begin{table}[h]
\renewcommand{\arraystretch}{1}
\centering
\tiny
\resizebox{1\columnwidth}{!}{
\begin{tabular}{@{}lcc@{}}
\toprule
Models/Datasets  & \begin{tabular}[c]{@{}c@{}}RoBERTa\\ AMR-LDA\end{tabular} & \begin{tabular}[c]{@{}c@{}}RoBERTa\\ LReasoner-LDA\end{tabular} \\ \midrule
Depth=1                  & 100.00                 & 100.00                       \\
Depth=1 (with altered rules)    & \textbf{100.00}                 & 99.87                  \\
Depth=2                  & 100.00                 & 100.00                       \\
Depth=2 (with altered rules)    & \textbf{99.73}            & 74.00                  \\ \bottomrule
\end{tabular}}
\caption{Comparison between AMR-LDA and LReasoner-LDA with RoBERTa-Large on PARARULE-Plus and PARARULE-Plus (with altered rules). Depth=1 means that only one rule was used to infer the answer. Depth=1 (with altered rules) means one of the rules has been altered using logical equivalence law.}
\label{comparison-multistep-deductive-reasoning-study}
\end{table}

We assessed the robustness of AMR-LDA and LReasoner-LDA models on the PARARULE-Plus dataset~\cite{bao2022multi} by modifying the test set with the contraposition law. Examples from this dataset can be found in \nameref{sec:appendix} Figures~\ref{PARARULE-Plus-depth-1} and \ref{PARARULE-Plus-depth-2}. AMR-LDA showed enhanced robustness on these altered tests compared to LReasoner-LDA.


\begin{table}[h]
\centering
\renewcommand{\arraystretch}{1}
\footnotesize
\resizebox{1\columnwidth}{!}{
\begin{tabular}{@{}lcccc@{}}
\toprule
Models/Datasets & \multicolumn{1}{c}{Con} & \multicolumn{1}{c}{Con-dou} & \begin{tabular}[c]{@{}c@{}}Con-dou\\ imp\end{tabular} & \begin{tabular}[c]{@{}c@{}}Con-dou\\ imp-com\end{tabular} \\ \midrule
\multicolumn{5}{c}{\emph{RoBERTa-Large as backbone model}} \\
ReClor         & 60.40                    & 60.80                    & \textbf{61.80}                    & 59.80                    \\
LogiQA         & 37.78                   & 33.17                   & 33.94                   & \textbf{38.70}           \\
MNLI           & 89.55                   &  \textbf{90.15}    & 89.68                   & 89.78          \\
MRPC           & 90.69                   & 89.22                   & 90.44                   & \textbf{90.93}          \\
RTE            & 81.23                   & 85.20                    & 84.84                   & \textbf{86.64}          \\
QNLI           & 94.16                   & 94.05                   & \textbf{94.51}                   & 94.49                   \\
QQP            & 92.12  & 89.88                   &  92.06    & \textbf{93.14}  \\ \midrule
\multicolumn{5}{c}{\emph{DeBERTaV2-XXLarge as backbone model}} \\
ReClor         & \textbf{81.80}
                    & 72.20
                    & 79.40                    & 78.80                    \\
LogiQA         & 32.25
                   & \textbf{45.46}
                   & 38.24
                   & 40.55 \\ \midrule
\multicolumn{5}{c}{\emph{DeBERTa-Large as backbone model}}          \\
MNLI           & \textbf{90.80}                   &  90.59    & 90.68                   & 89.67          \\
MRPC           & \textbf{90.20}                   & 88.48                   & 89.95                   & \textbf{90.20}          \\
RTE            & 84.84
                   & 87.36
                   & 85.56
                   & \textbf{88.09}          \\
QNLI           & \textbf{95.28}
                   & 95.04
                   & 94.97
                   & 95.24                   \\
QQP            & 92.33
  & 92.40
                   &  92.29
    & \textbf{92.47}        
\\ \bottomrule
\end{tabular}}
\caption{An experiment to assess the influence of different logical equivalence laws on downstream logical reasoning and natural language inference tasks. ``Con'', ``dou'', ``imp'' and ``com'' are the abbreviation for contraposition law, double negation law, implication law and commutative law. ``Con-dou'' denotes data constructed using both the contraposition law and the double negation law. Other terms are derived in a similar manner.}
\label{different-logical-equivalence-law-ablation-study}
\end{table}

\begin{table}[h]
\centering
\renewcommand{\arraystretch}{1}
\resizebox{\columnwidth}{!}{
\begin{tabular}{@{}lcccccc@{}}
\toprule
\multirow{2}{*}{Models/Datasets} & \multicolumn{4}{c}{ReClor}                                                                             & \multicolumn{2}{c}{LogiQA}        \\ \cmidrule(l){2-7} 
                                 & Dev             & Test                       & Test-E                     & Test-H                     & Dev             & Test            \\ \midrule
DeBERTaV2-XXLarge                      &  73.93         & 70.46 & 80.82 & 62.31 &   39.72        &   39.62                                                                                                                       \\
+ AMR-LDA-1:1                      & 78.80          & 76.10 & \textbf{84.77} & 69.28 & 40.55          & 41.47          \\
+ AMR-LDA-1:2                      & 80.20          & \textbf{76.40}            & \textbf{84.77}                     & \textbf{69.82}            & \textbf{47.00} & \textbf{43.93} \\
+ AMR-LDA-1:3                      & \textbf{81.20} & 75.70                     & 84.09                     & 69.10                     & 42.70          & 41.01          \\\midrule
DeBERTaV2-XXLarge + MERIt-1:3                        & 80.20          & 75.80                     & 85.00            & 68.57                     &  37.32               &  42.39               \\
 + AMR-LDA-Con-1:3                      & \textbf{82.60}          & 76.60 & 86.13 & \textbf{69.10} &     \textbf{45.00}      &      43.01     \\
 + AMR-LDA-Merged-1:3                      & 81.80          & \textbf{76.90}            &      \textbf{87.50}                &   68.57          & 44.54 & \textbf{45.62} \\
 \midrule
 DeBERTaV2-XXLarge + IDoL                        &   77.60        &       74.50               &   82.95         &      67.85                &        39.78         &     40.24            \\
+ AMR-LDA-Con-1:3                      &   79.20        & \textbf{77.00} & 85.68 & \textbf{70.17} &     \textbf{47.61}      &   \textbf{44.54}      \\
+ AMR-LDA-Merged-1:3                      &   \textbf{79.40}        &    75.60         &       \textbf{86.36}               &   67.14          & 41.93 & 41.32 \\
\bottomrule
\end{tabular}}
\caption{An experiment to assess how positive:negative sample ratios affect downstream tasks. AMR-LDA 1:1 means the ratio of positive and negative samples is 1:1.}
\label{pos-neg-ablation-study}
\end{table}


\subsection{Ablation Studies}
We perform experiments using a subset of the logical equivalence laws. We present the results in Table \ref{different-logical-equivalence-law-ablation-study}. This ablation study serves as the basis for our selection of four logical equivalence rules in the main experiment as Table~\ref{main-synthetic-test} shown. Since the test sets are private and used to rank models on the leaderboard, we evaluated directly using the validation sets instead of the test sets. To make a fair comparison, we ensure the sizes of the training sets are the same for con, con-dou, con-dou-imp and com-dou-imp-com. For this ablation study, we constructed training sets of size 1,000.

We conduct another ablation study where we modify the positive and negative sample ratios. We select DeBERTaV2-XXLarge as the backbone model. We compare the generated data against our AMR-LDA and MERIt. Table~\ref{pos-neg-ablation-study} shows that a higher proportion of negative samples may help increase the performance on logical reasoning tasks. Furthermore, we chose DeBERTaV2-XXLarge + MERIt-1:3~\cite{jiao2022merit} and DeBERTaV2-XXLarge + IDoL~\cite{xu-etal-2023-idol} as the backbone models. We performed logical equivalence identification contrastive learning, using data constructed solely from the AMR-LDA contraposition law and subsequently merging all four logical equivalence laws. Subsequent fine-tuning on downstream tasks demonstrated that incorporating more logical equivalence laws can enhance the performance of language models on logical reasoning tasks.

\section{Conclusion}
The sparsity of web data related to logical reasoning constrains the advancement of large language models in their performance on logical reasoning tasks. Existing methods for constructing logically equivalent sentences had been restricted to templates and specific datasets. Our AMR-LDA considers more logical equivalence laws than existing methods do, and it does not reply on any ad-hoc templates. We applied AMR-LDA to fine-tuning discriminative LLMs and prompt augmentation of generative LLMs (GPT-3.5 and GPT-4), yielding better results than baseline methods on logical reasoning tasks.

\section{Human Evaluation}
Human evaluation was conducted to evaluate the correctness and fluency of the logically manipulated sentences generated using AMR-LDA and LReasoner-LDA. We constructed a survey with 20 questions,  each question consisting of two randomly selected sentences: one from those generated by our AMR-LDA and the other by LReasoner-LDA. 45 participants completed the survey anonymously. We asked them to evaluate the sentences in two aspects: 1) which sentence is logically equivalent to the original sentence, or whether both of them are logically equivalent to the original sentence, and 2) which sentence is more fluent. 63.92\% and 76.44\% of people prefered AMR-LDA's logically equivalent and fluent sentences over those generated by LReasoner-LDA.


\section{Limitations}

One limitation of our approach is its reliance on AMR for logic-driven data augmentation, which, while innovative, may not fully capture the intricacies of natural language variation and complex logical constructs encountered in diverse texts. This constraint reflects the broader challenge in NLP of developing models that can understand and reason with the full spectrum of human language, including idiomatic expressions, nuanced context, and varied logical frameworks. Our work makes significant strides in this direction, yet it also highlights the need for continued research to enhance the robustness and adaptability of NLP systems to more closely mirror human-level comprehension and reasoning capabilities. 


\section{Ethics Statement}
All the data used in this paper are either synthetically generated or open-source datasets. All the code used to run the experiments is written using open-source libraries or adapted from published code from other papers. We will also release our code and any synthetically generated data to ensure that the work can be reproduced. The human evaluation was approved by the Ethics Committee of the main authors' employer. 

\section{Future Work}
It is worth exploring how data augmentation can be used for dynamic prompt tuning in logical reasoning tasks~\cite{10651072,qi2023a,10.1145/3373017.3373049}. Several studies~\cite{bao2023assessing,10174688} have explored task variation formats of ReClor, LogiQA, and LogiQA-2 by altering the order of options or replacing the answers, and have found that large language models perform significantly worse under these variations. It is also worth exploring how AMR can work in conjunction with logic programming to iteratively improve reasoning performance~\cite{10650138,baosymbolic,doi:10.1142/S2705078521500156,tan-etal-2023-multi2claim,Ni_Bao_Li_Qi_Denny_Warren_Witbrock_Liu_2022,bao2022natural,Bao_2025,bao2025developing,gendron2023large}. Furthermore, it is worth investigating how alternative LoRA fine-tuning methods can be used to train only the LoRA adapters~\cite{xiao2024cora}.

\bibliography{acl_latex}

\appendix

\section{Appendix}
\label{sec:appendix}

\section{Experiment Setup}\label{experiment_setup}
We follow the training script from Huggingface and the default hyperparameters\footnote{\url{https://github.com/huggingface/transformers/tree/main/examples/pytorch/text-classification}} to conduct the training and Algorithms~\ref{negative-sample-construction} and~\ref{representation-learning} illustrate the negative sample construction and the training process, respectively. For the contrastive learning, we fine-tune RoBERTa-Large, DeBERTa-Large, and DeBERTaV2-XXLarge using the constructed logical equivalence sentence pair from our AMR-LDA, LReasoner's logic-driven data augmentation method (LReasoner-LDA) and AMR-DA data augmentation method. We use DeBERTaV2-XXLarge for ReClor and LogiQA tasks because DeBERTaV2 supports multiple-choice question tasks with a DeBERTaV2ForMultipleChoice head. The hyperparameters for stages 1 and 2 training can be found in Tables~\ref{hyperparameter-details-stage1-stage2} and~\ref{hyperparameter-reclor-logiqa-stage2}.  

\section{Conversion Between Texts and AMR}
\label{preliminary_study}
In order to decide which models to use to perform text and AMR conversions, we experiment with different combinations of text-to-AMR and AMR-to-text models. In the experiment, a sentence is converted to AMR, and then is converted back to text without any modification to the AMR. We pick the combination that can recover the original sentence the most, as measured in BLEU score. The results are reported in Table~\ref{amr-model-selection}. We find that using parse\_xfm\_bart large as the AMR parser and T5Wtense as the AMR generator produces the highest BLEU score. Therefore, we select them as the text-to-AMR parser and AMR-to-text generator in all the remaining experiments. Parse\_xfm\_bart\_large is an AMR parser that uses BART-Large as the backbone model~\cite{lewis2019bart}. T5Wtense is an AMR generator that uses T5 as the backbone model~\cite{raffel2020exploring}.

\begin{table}[h]
\renewcommand{\arraystretch}{1}
\centering
\setlength{\tabcolsep}{20pt}
\resizebox{1\columnwidth}{!}{
\begin{tabular}{@{}lcc@{}}
\toprule
Text-To-AMR Parser                       & AMR-To-Text Generator & BLEU           \\ \midrule
                                         & Spring                & 25.08          \\
Spring                                   & T5wtense              & 30.86          \\
                                         & T5                    & 24.76          \\ \midrule
\multirow{2}{*}{T5}                      & T5wtense              & 29.33          \\
                                         & T5                    & 30.82          \\ \midrule
\multirow{2}{*}{parse\_xfm\_bart\_large} & T5wtense              & \textbf{38.45} \\
                                         & T5                    & 30.10          \\ \bottomrule
\end{tabular}}
\caption{Comparison of different combinations of text-to-AMR and AMR-to-text models in recovering original texts after the conversions without any augmentation to the AMR. We adopt the combination with the highest BLEU score in the rest of the experiments.}
\label{amr-model-selection}
\end{table}

\section{Case Studies}\label{case-studies}
We present several case studies comparing our AMR-LDA method with LReasoner-LDA in terms of constructing logically equivalent sentences. These constructions leverage four logical equivalence laws. LReasoner-LDA, however, does not design for the implication law, double negation law, or the commutative law, leading to its inability to handle scenarios that require these laws. Additionally, LReasoner-LDA struggles to construct logically equivalent sentences using the contraposition law when encountering new sentences not found in the ReClor and LogiQA datasets.

\begin{table}[H]
\renewcommand{\arraystretch}{0.6}
\centering
\setlength{\tabcolsep}{20pt}
\resizebox{\columnwidth}{!}{
\begin{tabular}{@{}ll@{}}
\toprule
                                   & \multicolumn{1}{c}{Contraposition law}                  \\ \midrule
\multirow{2}{*}{Original Sentence} & If the bald eagle is small,         \\
                                   & then the mouse is not small.        \\
\multirow{2}{*}{AMR-LDA}            & The bald eagle isn't small,         \\
                                   & unless the mouse is small.          \\
\multirow{2}{*}{LReasoner-LDA}      & If it is not small, then it                 \\
                                   & will be not the bald eagle. \\ \bottomrule
\end{tabular}}
\caption{Logically equivalent sentences constructed by contraposition law.}
\label{case-study-1}
\end{table}

\begin{table}[H]
\renewcommand{\arraystretch}{0.6}
\centering
\setlength{\tabcolsep}{20pt}
\resizebox{\columnwidth}{!}{
\begin{tabular}{@{}ll@{}}
\toprule
                                   & \multicolumn{1}{c}{Contraposition law}                  \\ \midrule
\multirow{2}{*}{Original Sentence} & If the bald eagle is kind,          \\
                                   & then Dave is not short.             \\
\multirow{2}{*}{AMR-LDA}            & If Dave is short,                   \\
                                   & the bald eagle is not kind.         \\
\multirow{2}{*}{LReasoner-LDA}      & If it is not kind, then it                 \\
                                   & will be not the bald eagle. \\ \bottomrule
\end{tabular}}
\caption{Logically equivalent sentences constructed by contraposition law.}
\label{case-study-2}
\end{table}

\begin{table}[H]
\renewcommand{\arraystretch}{0.6}
\centering
\setlength{\tabcolsep}{20pt}
\resizebox{\columnwidth}{!}{
\begin{tabular}{@{}ll@{}}
\toprule
                                   & \multicolumn{1}{c}{Implication law}                        \\ \midrule
\multirow{2}{*}{Original Sentence} & The bear is not sleepy                 \\
                                   & or Bob is not cute.                    \\
\multirow{2}{*}{AMR-LDA}            & If the bear is sleepy,             \\
                                   & then Bob is not cute.                  \\
\multirow{2}{*}{LReasoner-LDA}      & \multicolumn{1}{c}{\multirow{2}{*}{-}} \\
                                   & \multicolumn{1}{c}{}                   \\ \bottomrule
\end{tabular}}
\caption{Logically equivalent sentences constructed by implication law.}
\label{case-study-3}
\end{table}

\begin{table}[H]
\renewcommand{\arraystretch}{0.6}
\centering
\setlength{\tabcolsep}{20pt}
\resizebox{\columnwidth}{!}{
\begin{tabular}{@{}ll@{}}
\toprule
                  & \multicolumn{1}{c}{Double negation law}           \\ \midrule
Original Sentence & The bald eagle is beautiful.

       \\
AMR-LDA            & The bald eagle isn't ugly.

    \\
LReasoner-LDA         & \multicolumn{1}{c}{\multirow{1}{*}{-}}
 \\ \bottomrule
\end{tabular}}
\caption{Logically equivalent sentences constructed by double negation law.}
\label{case-study-4}
\end{table}

\begin{table}[H]
\renewcommand{\arraystretch}{0.6}
\centering
\setlength{\tabcolsep}{20pt}
\resizebox{\columnwidth}{!}{
\begin{tabular}{@{}ll@{}}
\toprule
                                   & \multicolumn{1}{c}{Implication law}                        \\ \midrule
\multirow{2}{*}{Original Sentence} & If the lion is not funny,              \\
                                   & then the tiger is beautiful.           \\
\multirow{2}{*}{AMR-LDA}            & The lion is funny                     \\
                                   & or the tiger is beautiful.                   \\
\multirow{2}{*}{LReasoner-LDA}      & \multicolumn{1}{c}{\multirow{2}{*}{-}} \\
                                   & \multicolumn{1}{c}{}                   \\ \bottomrule
\end{tabular}}
\caption{Logically equivalent sentences constructed by implication law.}
\label{case-study-5}
\end{table}

\begin{table}[H]
\renewcommand{\arraystretch}{0.6}
\centering
\setlength{\tabcolsep}{20pt}
\resizebox{\columnwidth}{!}{
\begin{tabular}{@{}ll@{}}
\toprule
                  & \multicolumn{1}{c}{Double negation law}           \\ \midrule
Original Sentence & The bald eagle is strong.

       \\
AMR-LDA            & The bald eagle is not weak.

    \\
LReasoner-LDA         & \multicolumn{1}{c}{\multirow{0.6}{*}{-}}
 \\ \bottomrule
\end{tabular}}
\caption{Logically equivalent sentences constructed by double negation law.}
\label{case-study-6}
\end{table}

\begin{table}[H]
\renewcommand{\arraystretch}{0.6}
\centering
\setlength{\tabcolsep}{20pt}
\resizebox{\columnwidth}{!}{
\begin{tabular}{@{}ll@{}}
\toprule
                                   & \multicolumn{1}{c}{Commutative law}                        \\ \midrule
\multirow{2}{*}{Original Sentence} & The bald eagle is kind                 \\
                                   & and the wolf is not dull.              \\
\multirow{2}{*}{AMR-LDA}            & The wolf is not dull                   \\
                                   & and the bald eagle is kind.            \\
\multirow{2}{*}{LReasoner-LDA}      & \multicolumn{1}{c}{\multirow{2}{*}{-}} \\
                                   & \multicolumn{1}{c}{}                   \\ \bottomrule
\end{tabular}}
\caption{Logically equivalent sentences constructed by commutative law.}
\label{case-study-7}
\end{table}

\begin{table}[H]
\centering
\renewcommand{\arraystretch}{0.6}
\setlength{\tabcolsep}{20pt}
\resizebox{\columnwidth}{!}{
\begin{tabular}{@{}ll@{}}
\toprule
                                   & \multicolumn{1}{c}{Commutative law}                        \\ \midrule
\multirow{2}{*}{Original Sentence} & The lion is thin                       \\
                                   & and the dinosaur is not angry.         \\
\multirow{2}{*}{AMR-LDA}            & The dinosaur was not angry             \\
                                   & and the lion was thin.                 \\
\multirow{2}{*}{LReasoner-LDA}      & \multicolumn{1}{c}{\multirow{2}{*}{-}} \\
                                   & \multicolumn{1}{c}{}                   \\ \bottomrule
\end{tabular}}
\caption{Logically equivalent sentences constructed by commutative law.}
\label{case-study-8}
\end{table}

\subsection{Real World/Long Sentence Case Studies}\label{long-sentence-example}

\begin{figure*}[h]
\centering
\begin{tcolorbox}[title=AMR-LDA Prompt Augmentation Case Study]
            \textbf{GPT-4 Input:} ``context'': ``If you have no keyboarding skills at all, you will not be able to use a computer. And if you are not able to use a computer, you will not be able to write your essays using a word processing program.'', 

            ``question'': ``If the statements above are true, which one of the following must be true?'', "answers": 

			A. ``If you are not able to write your essays using a word processing program, you have no keyboarding skills. \textbf{\textit{\textcolor{blue}{If you have the skill of a keyboard, you can write your essay using a word processing program.}\textcolor{cyan}{If you can use a computer, you have keyboarding skills. If you can write your essay with a word processing program, you can use a computer.} \textcolor{brown}{Whether you have keyboard skills at all or can't use a computer. Whether you can use a computer or you can't write your own essay with a word processing program.}}}'',

            B. ``If you are able to write your essays using a word processing program, you have at least some keyboarding skills. \textbf{\textit{\textcolor{blue}{If you don't have at least some keyboard skills, you can't write your essay with a word processing program.} \textcolor{cyan}{If you can use a computer, you have keyboarding skills. If you can write your essay with a word processing program, you can use a computer.} \textcolor{brown}{Whether you have keyboard skills at all or can't use a computer. Whether you can use a computer or you can't write your own essay with a word processing program.}}}'',
            
			C. ``If you are not able to write your essays using a word processing program, you are not able to use a computer. \textbf{\textit{\textcolor{blue}{If you can use a computer, you can write your essay using word processing programs.} \textcolor{cyan}{If you can use a computer, you have keyboarding skills. If you can write your essay with a word processing program, you can use a computer.} \textcolor{brown}{Whether you have keyboard skills at all or can't use a computer. Whether you can use a computer or you can't write your own essay with a word processing program.}}}'',
   
			D. ``If you have some keyboarding skills, you will be able to write your essays using a word processing program. \textbf{\textit{\textcolor{blue}{If you can't write your essay with a word processing program, you don't have some keyboard skills.} \textcolor{cyan}{If you can use a computer, you have keyboarding skills. If you can write your essay with a word processing program, you can use a computer.} \textcolor{brown}{Whether you have keyboard skills at all or can't use a computer. Whether you can use a computer or you can't write your own essay with a word processing program.}}}''

\textbf{GPT-4 output: B}
\end{tcolorbox}
\caption{Example for using AMR-LDA to augment the prompt from ReClor dataset and their subsequent utilisation as input for GPT-4. Data segments that are marked in bold italics and appear in blue were generated using the contraposition law, while those in brown were generated using the implication law.}
\label{fig:amr-lda-prompt-augmentation}
\end{figure*}

The appendix of our paper describes Algorithm~\ref{leamr-data-augmentation}, which uses four lists from Tables~\ref{logic-pattern-double-negation},~\ref{logic-pattern-commutative-law},~\ref{logic-pattern-contraposition-law} and~\ref{logic-pattern-implication-law} to create synthetic sentences. We've also tested our method on real-world datasets like ReClor and LogiQA that require logical reasoning. Our method, AMR-LDA prompt augmentation, can work with just one list of various sentences. It automatically detects if a sentence can be transformed into a logically equivalent one using a specific logical equivalence law. An example of this application on a real-world sentence is shown in Figure~\ref{fig:amr-lda-prompt-augmentation}. We process sentences from context and options, generating logically equivalent sentences where possible.

Our AMR-LDA can also been applied to long sentences. Our method can generate logically equivalent sentences for long sentences with clear sentence structure using logical equivalence rules (Commutative law) as shown in Figure~\ref{fig:long-example-1} and~\ref{fig:long-example-2}. The second example shows that our AMR-LDA can understand the effect of that clause on yoga stretching, showing the generalisation advantages of AMR as a semantic representation compared to LReasoner-LDA which relies on a constituency parser and template and fails in this case which is out of templates. 

\begin{figure*}[]
\centering
\begin{tcolorbox}[title=Long Sentence Example 1:]

\textbf{Original sentence:} Sarah woke up early in the morning, and she started her day with a cup of coffee and some light yoga stretches.

\textbf{Original sentence's AMR graph:} (a / and
      :op1 \textcolor{blue}{(w / wake-up-02
	        :ARG1 (p / person
	              :name (n / name
	              :op1 "Sarah"))
	        :time (e / early
	              :op1 (d / date-entity
	              :dayperiod (m / morning))))}
      :op2 \textcolor{brown}{(s / start-01
	        :ARG0 p
	        :ARG1 (d2 / day
	              :poss p)
	        :ARG2 (a2 / and
	              :op1 (c / coffee
	              :quant (v / volume-quantity
	                    :quant 1
	                    :unit (c2 / cup)))
	              :op2 (s2 / stretch-01
	                    :ARG0 p
	                    :mod (y / yoga)
	                    :ARG1-of (l / light-06)
	                    :quant (s3 / some)))))}

\textbf{Modified AMR graph using AMR-LDA:} (a / and
      :op1 \textcolor{brown}{(s / start-01
            :ARG0 p
            :ARG1 (d2 / day
                  :poss p)
            :ARG2 (a2 / and
                  :op1 (c / coffee
                        :quant (v / volume-quantity
                              :quant 1
                              :unit (c2 / cup)))
                  :op2 (s2 / stretch-01
                        :ARG0 p
                        :mod (y / yoga)
                        :ARG1-of (l / light-06)
                        :quant (s3 / some))))}
      :op2 \textcolor{blue}{(w / wake-up-02
            :ARG1 (p / person
                  :name (n / name
                        :op1 "Sarah"))
            :time (e / early
                  :op1 (d / date-entity
                        :dayperiod (m / morning)))))}

\textbf{Generated logical equivalence sentence using AMR-LDA:} Sarah started her day with a cup of coffee and some light yoga stretching and woke up early in the morning.

\end{tcolorbox}
\caption{One example uses our AMR-LDA to generate logical equivalence sentences for long sentences. In this case, a logical equivalence sentence is generated using commutative law, and the same color represents the same argument. In this case, the order of the former and latter arguments for the conjunction word ``and'' has been swapped.}
\label{fig:long-example-1}
\end{figure*}

\begin{figure*}[]
\centering
\begin{tcolorbox}[title=Long Sentence Example 2:]

\textbf{Original sentence:} Sarah woke up early in the morning, and she started her day with a cup of coffee and some light yoga stretches that will help lose weight.

\textbf{Original sentence's AMR graph:} (a / and
      (a / and
      :op1 \textcolor{blue}{(w / wake-up-02
	  	  :ARG1 (p / person
	  	  	  :name (n / name
	  	  	  	  :op1 "Sarah"))
	  	  :time (e / early
	  	  	  :op1 (d / date-entity
	  	  	  	  :dayperiod (m / morning))))}
      :op2 \textcolor{brown}{(s / start-01
	  	  :ARG0 p
	  	  :ARG1 (d2 / day
	  	  	  :poss p)
	  	  :ARG2 (a2 / and
	  	  	  :op1 (c / coffee
	  	  	  	  :quant (v / volume-quantity
	  	  	  	  	  :quant 1
	  	  	  	  	  :unit (c2 / cup)))
	  	  	  :op2 (s2 / stretch-01
	  	  	  	  :mod (y / yoga)
	  	  	  	  :ARG0-of (h / help-01
	  	  	  	  	  :ARG1 (l / lose-01
	  	  	  	  	  	  :ARG1 (w2 / weight)))
	  	  	  	  :ARG1-of (l2 / light-06)
	  	  	  	  :quant (s3 / some)))))}

\textbf{Modified AMR graph using AMR-LDA:} (a / and
      :op1 \textcolor{brown}{(s / start-01
            :ARG0 p
            :ARG1 (d2 / day
                  :poss p)
            :ARG2 (a2 / and
                  :op1 (c / coffee
                        :quant (v / volume-quantity
                              :quant 1
                              :unit (c2 / cup)))
                  :op2 (s2 / stretch-01
                        :mod (y / yoga)
                        :ARG0-of (h / help-01
                              :ARG1 (l / lose-01
                                    :ARG1 (w2 / weight)))
                        :ARG1-of (l2 / light-06)
                        :quant (s3 / some))))}
      :op2 \textcolor{blue}{(w / wake-up-02
            :ARG1 (p / person
                  :name (n / name
                  :op1 "Sarah"))
            :time (e / early
                  :op1 (d / date-entity
                  :dayperiod (m / morning)))))}

\textbf{Generated logical equivalence sentence using AMR-LDA:} Sarah started her day with a cup of coffee and some light yoga stretching to help lose weight, and woke up early in the morning.

\end{tcolorbox}
\caption{One example uses our AMR-LDA to generate logical equivalence sentences for long sentences. In this case, a logical equivalence sentence is generated using commutative law, and the same color represents the same argument. AMR-LDA can understand the effect of that clause on yoga stretching. In this case, the order of the former and latter arguments for the conjunction word ``and'' has been swapped.}
\label{fig:long-example-2}
\end{figure*}
\section{Synthetic Dataset Construction}\label{synthetic-data-construction}
Here are the entities, relationships, and attributes we used to construct our synthetic dataset. We used the synthetic dataset to conduct the AMR-based logic-driven data augmentation and logical-equivalence-identification contrastive learning. For the subject, we used ``the bald eagle'', ``the tiger'', ``the bear'', ``the lion'', ``the wolf'', ``the crocodile'', ``the dinosaur'', ``the snake'', ``the leopard'', ``the cat'', ``the dog'', ``the mouse'', ``the rabbit'', ``the squirrel'', ``Anne'', ``Alan'', ``Bob'', ``Charlie'', ``Dave'', ``Erin'', ``Harry'', ``Gary'', and ``Fiona''. For the relationships, we used ``is'' and ``is not''. For the attributes, we used ``kind'', ``quiet'', ``round'', ``nice'', ``smart'', ``clever'', ``dull'', ``rough'', ``lazy'', ``slow'', ``sleepy'', ``boring'', ``tired'', ``reckless'', ``furry'', ``small'', ``cute'', ``lovely'', ``beautiful'', ``funny'', ``big'', ``strong'', ``awful'', ``fierce'', ``heavy'', ``horrible'', ``powerful'', ``angry'', ``tall'', ``huge'', ``short'', ``thin'', ``little'', ``tiny'', ``wealthy'', ``poor'', ``dull'', ``rough'', ``bad'', and ``sad''. 

Here are the entities, relationships, and attributes we used to fine-tune T5-Large. After T5-Large had been fine-tuned, we used the fine-tuned model to generate logical equivalence sentences as the label for the above synthetic sentences and then conducted the logical-equivalence-identification contrastive learning and downstream task. For the subject, based on the above subject name entities, we add ``the duck'', ``the goat'', ``the goose'', ``the donkey'', ``the cow'', ``James'', ``Robert'', ``John'', ``Michael'', ``David'', ``William'', ``Richard'', ``Anthony'', ``Paul'', ``Andrew''. For the attributes, we add ``cautious'', ``careful'', ``brainy'', ``bored'', ``adorable'', ``aggressive'', ``anxious'', ``dizzy'', ``depressed'', ``disturbed'', and ``awful''. 

\begin{table*}[h]
\centering
\begin{tabular}{@{}lc@{}}
\toprule
                  & Logic pattern for double negation law                                \\ \midrule
Original sentence & subject + verb + adj                            \\
Positive sample   & subject + verb + ``not'' + the antonym of the adj \\
Negative sample   & subject + verb + ``not'' + adj                    \\ \bottomrule
\end{tabular}
\caption{We used the logic pattern for double negation law for constructing the test set for the experiment in Table \ref{preliminary-synthetic-test}.}
\label{logic-pattern-double-negation}
\end{table*}

\begin{table*}[h]
\centering
\begin{tabular}{@{}lcc@{}}
\toprule
                  & Original logic pattern for commutative law & Changed logic pattern \\ \midrule
s1                & sub1 + verb1 +  adj1                       & sub1 +  verb1 + ``not'' + adj1              \\
s2                & sub2 + verb2 +  adj2                       & sub2 + verb2 + ``not'' + adj2               \\
s3                & sub1 + verb1 + ``not'' + adj1                & sub2 + verb2 + ``not'' + adj2               \\
Original sentence & \multicolumn{2}{c}{s1 + ``and'' + s2}                                                    \\
Positive sample   & \multicolumn{2}{c}{s2 + ``and'' + s1}                                                    \\
Negative sample   & \multicolumn{2}{c}{s1 + ``and'' + s3}                                                    \\ \bottomrule
\end{tabular}
\caption{We used the logic pattern for commutative law for constructing the test set for the experiment in Table \ref{preliminary-synthetic-test}.}
\label{logic-pattern-commutative-law}
\end{table*}

\begin{table*}[h]
\centering
\begin{tabular}{@{}lc@{}}
\toprule
                   & Logic pattern for contraposition law                                       \\ \midrule
Original sentence1 & ``If'' + sub1 + verb + adj1 + ``, then'' + sub2 + verb + adj2                  \\
Positive sentence1 & ``If'' + sub2 + verb + ``not'' + adj2 + ``, then'' + sub1 + verb + ``not'' + adj1  \\
Negative sentence1 & ``If'' + sub1 + verb + adj1 + ``, then'' + sub2 + verb + ``not'' + adj2          \\ \midrule
Original sentence2 & ``If'' + sub1 + verb + adj1 + ``, then'' + sub2 + verb + ``not'' + adj2          \\
Positive sentence2 & ``If'' + sub2 + verb + adj2 + ``, then'' + sub1 + verb + ``not'' + adj1          \\
Negative sentence2 & ``If'' + sub1 + verb + adj1 + ``, then'' + sub2 + verb + adj2                  \\ \midrule
Original sentence3 & ``If'' + sub1 +  verb + ``not'' + adj1 + ``, then'' + sub2 + verb + adj2         \\
Positive sentence3 & ``If'' + sub2 + verb + ``not'' + adj2 + ``, then'' + sub1 + verb + adj1          \\
Negative sentence3 & ``If'' + sub1 + verb + ``not'' + adj1 + ``, then'' + sub2 + verb + ``not'' + adj2 \\ \midrule
Original sentence4 & ``If'' + sub1 + verb + ``not'' + adj1 + ``, then'' + sub2 + verb + ``not'' + adj2  \\
Positive sentence4 & ``If'' + sub2 + verb + ``not'' + adj2 + ``, then'' + sub1 + verb + ``not'' + adj1  \\
Negative sentence4 & ``If'' + sub1 + verb + ``not'' + adj1 + ``, then'' + sub2 + verb + adj2          \\ \bottomrule
\end{tabular}
\caption{We used the logic pattern for contraposition law for constructing the test set for the experiment in Table \ref{preliminary-synthetic-test}.}
\label{logic-pattern-contraposition-law}
\end{table*}

\begin{table*}[h]
\centering
\begin{tabular}{@{}lc@{}}
\toprule
                  & Original logic pattern for implication law                     \\ \midrule
Original sentence & ``If'' + sub1 + verb + adj1 + ``, then'' + sub2 + verb + adj2      \\
Positive sample   & sub1 + verb + ``not'' + adj1 + ``or'' + sub2 + verb + adj2                 \\
Negative sample   & sub1 + verb + ``not'' + adj1 + ``or'' + sub2 + verb + ``not'' + adj2 \\ \midrule
                  & Changed logic pattern                                          \\
Original sentence & sub1 + verb + ``not'' + adj1 + ``or'' + sub2 + verb + adj2                 \\
Positive sample   & ``If'' + sub1 + verb + adj1 + ``, then'' + sub2 + verb + adj2      \\
Negative sample   & sub1 + verb + ``not'' + adj1 + ``or'' + sub2 + verb + ``not'' + adj2 \\ \bottomrule
\end{tabular}
\caption{We used the logic pattern for implication law for constructing the test set for the experiment in Table \ref{preliminary-synthetic-test}.}
\label{logic-pattern-implication-law}
\end{table*}

The entity names used for the ``change name'' experiment in Table \ref{preliminary-synthetic-test}. For the new entity names that we used ``the sheep'', ``the kitten'', ``the Garfield'', ``the lion'', ``the goat'', ``the bull'', ``the cow'', ``the elephant'', ``the butterfly'', ``the fish'', ``Peter'', ``Bill'', ``Tom'', ``Amy'', ``Charles'', ``Tim'', ``Lucy'', and ``John''.

Table \ref{logic-pattern-double-negation}, \ref{logic-pattern-commutative-law}, \ref{logic-pattern-contraposition-law}, and \ref{logic-pattern-implication-law} are the logic pattern and its variation that we consider to replace the original logic pattern for the experiment on Table \ref{preliminary-synthetic-test}.

\begin{table*}[h]
\renewcommand{\arraystretch}{1}
\centering
\setlength{\tabcolsep}{20pt}
\resizebox{0.6\textwidth}{!}{
\begin{tabular}{@{}lcc@{}}
\toprule
Test sets $\downarrow$; Models $\rightarrow$   & RoBERTa    & \begin{tabular}[c]{@{}c@{}}Fine-tuned\\ RoBERTa\end{tabular} \\ \midrule
Test set 1                        & 53.35 & 85.13        \\
Test set 2 (change name)          & 53.47 & 85.10        \\
Test set 3 (change logic) & 46.72 & 94.82        \\ \bottomrule
\end{tabular}}
\caption{Compared fine-tuned RoBERTa-Large and RoBERTa-Large on three different synthetic test sets.}
\label{preliminary-synthetic-test}
\end{table*}

To validate whether pre-trained language model can distinguish logically equivalent sentences. We design a preliminary experiment as Table~\ref{preliminary-synthetic-test} shown. We use RoBERTa-Large to conduct the experiment. We first generate a synthetic test set 1, which includes 1312 test samples with 23 entities, 2 relationships, 40 attributes, and 4 logical equivalence laws (double negation, contraposition, implication, and commutative laws). Model's performance can improve if we fine-tune language model on the logical equivalence training set, which is constructed by our AMR-LDA data augmentation method. Also, The result shows that the model's performance will not drop if we change the entity name or logic pattern, this indicates that the fine-tuned discriminative large language model can handle scenarios requiring greater robustness more effectively.

\begin{table*}[h]
\renewcommand{\arraystretch}{1}
\centering
\resizebox{0.55\textwidth}{!}{
\begin{tabular}{@{}lcc@{}}
\toprule
                             & \begin{tabular}[c]{@{}c@{}}Stage-1\\ Fine-tuning\end{tabular} & \begin{tabular}[c]{@{}c@{}}Stage-2\\ Fine-tuning\end{tabular} \\ \midrule
Seed                         & 2021               & 0/21/42                 \\
Batch Size                   & 32                 & 16/32              \\
Initial Learning Rate        & 2e-5               & 2e-5/3e-6          \\
Learning Rate Scheduler Type & \multicolumn{2}{c}{Linear}              \\
Epoch                        & \multicolumn{2}{c}{10}                  \\
Num Warmup Steps             & \multicolumn{2}{c}{0}                   \\
Weight Decay                 & \multicolumn{2}{c}{0}                   \\
Max Sequence Length          & \multicolumn{2}{c}{256}                 \\
Gradient Accumulation Steps  & \multicolumn{2}{c}{1}                   \\ \bottomrule
\end{tabular}}
\caption{Hyperparameter details for stage-1 fine-tuning and stage-2 fine-tuning except ReClor and LogiQA. Stage-1 fine-tuning means logical-equivalence-identification contrastive learning, and stage-2 fine-tuning means fine-tuning on the downstream tasks.}
\label{hyperparameter-details-stage1-stage2}
\end{table*}

\begin{table*}[h]
\renewcommand{\arraystretch}{1}
\scriptsize
\centering
\resizebox{0.55\textwidth}{!}{
\begin{tabular}{@{}lcc@{}}
\toprule
                            & \multicolumn{2}{c}{Stage-2 Fine-tuning} \\
                            & ReClor             & LogiQA            \\ \midrule
Seed & \multicolumn{2}{c}{42}   \\   
Batch Size                  & \multicolumn{2}{c}{2/4}                  \\
Gradient Accumulation Steps & \multicolumn{2}{c}{2}                  \\
Initial Learning Rate       & \multicolumn{2}{c}{1e-05/1e-5/3e-6}              \\
Epoch                       & \multicolumn{2}{c}{10}                 \\
Adam Betas                  & \multicolumn{2}{c}{(0.9, 0.98)}        \\
Adam Epsilon                & \multicolumn{2}{c}{1e-6}               \\
No Clip Grad Norm           & \multicolumn{2}{c}{True}               \\
Warmup Proportion           & \multicolumn{2}{c}{0.1}                \\
weight\_decay               & \multicolumn{2}{c}{0.01}               \\ \bottomrule
\end{tabular}}
\caption{Model hyperparameter tuning details for stage-2 fine-tuning on ReClor and LogiQA.}
\label{hyperparameter-reclor-logiqa-stage2}
\end{table*}

\begin{figure*}[h]
\centering
\includegraphics[width=500pt]{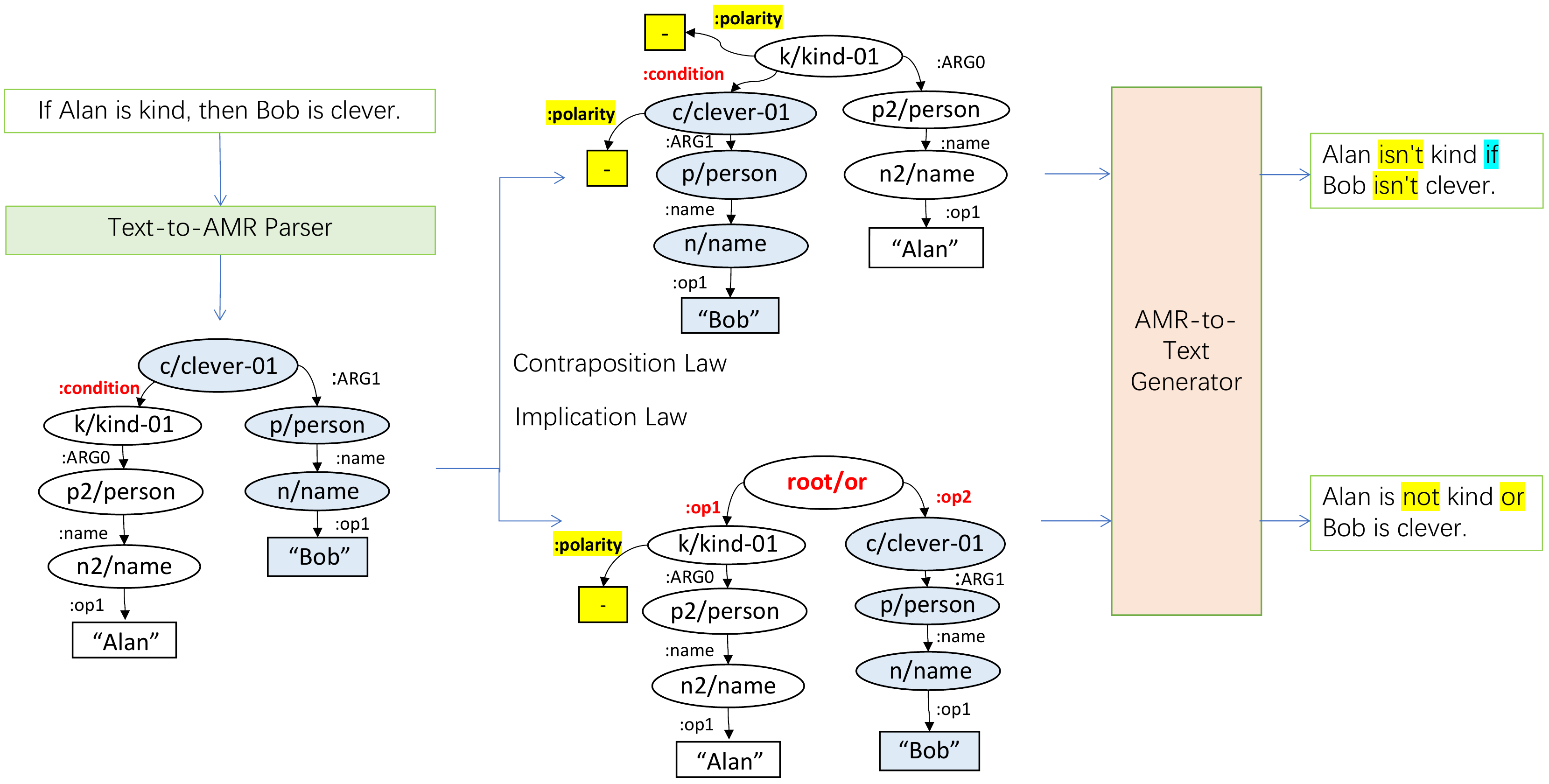}
\caption{An example of our AMR-based logic-driven data augmentation method using contraposition law and implication law}
\label{contraposition-implication-example}
\end{figure*}

\begin{figure*}[h]
\centering
\includegraphics[width=328pt]{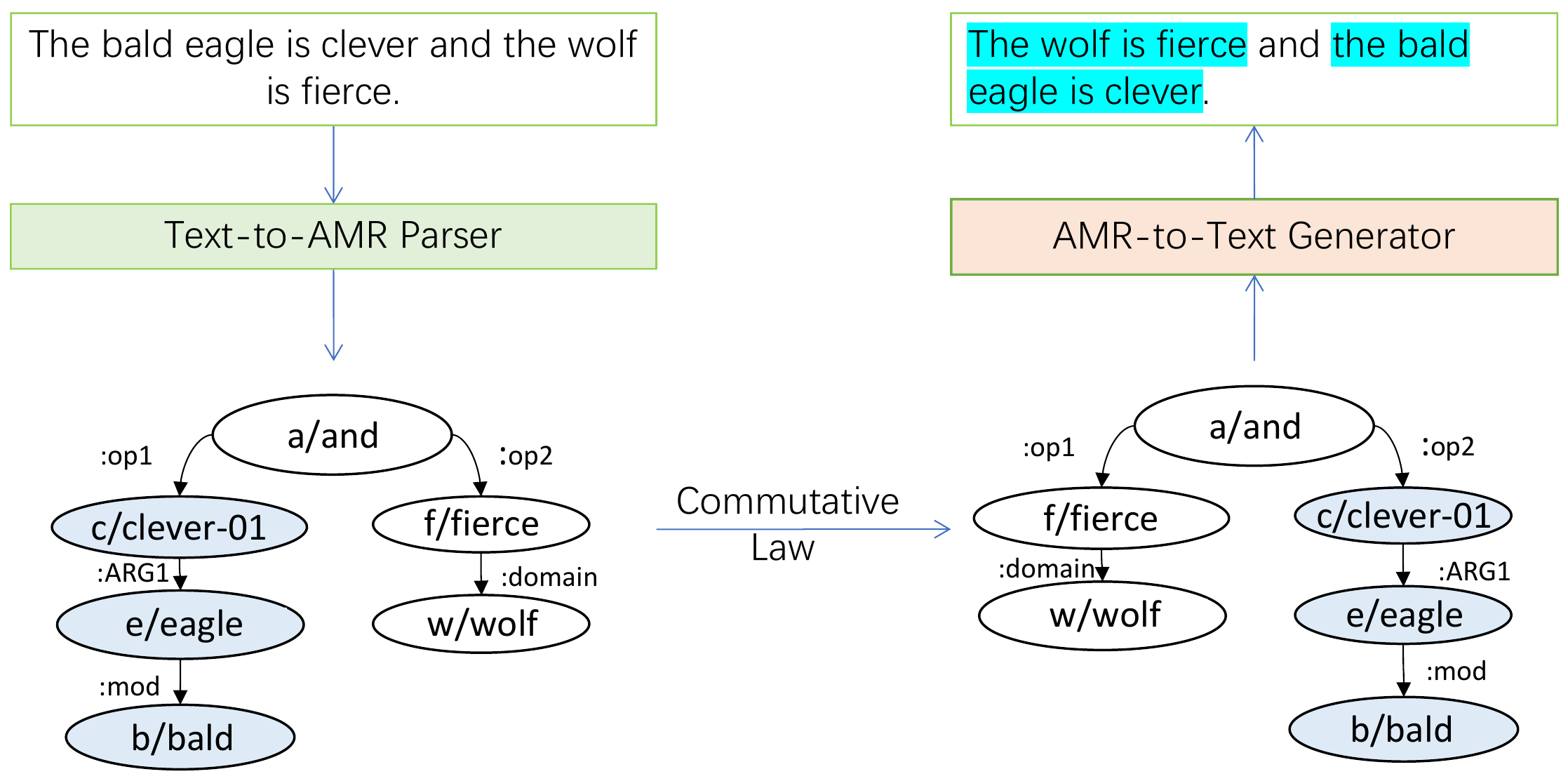}
\caption{An example of our AMR-based logic-driven data augmentation method using commutative law}
\label{commutative-example}
\end{figure*}

\begin{figure*}[h]
\centering
\includegraphics[width=357pt]{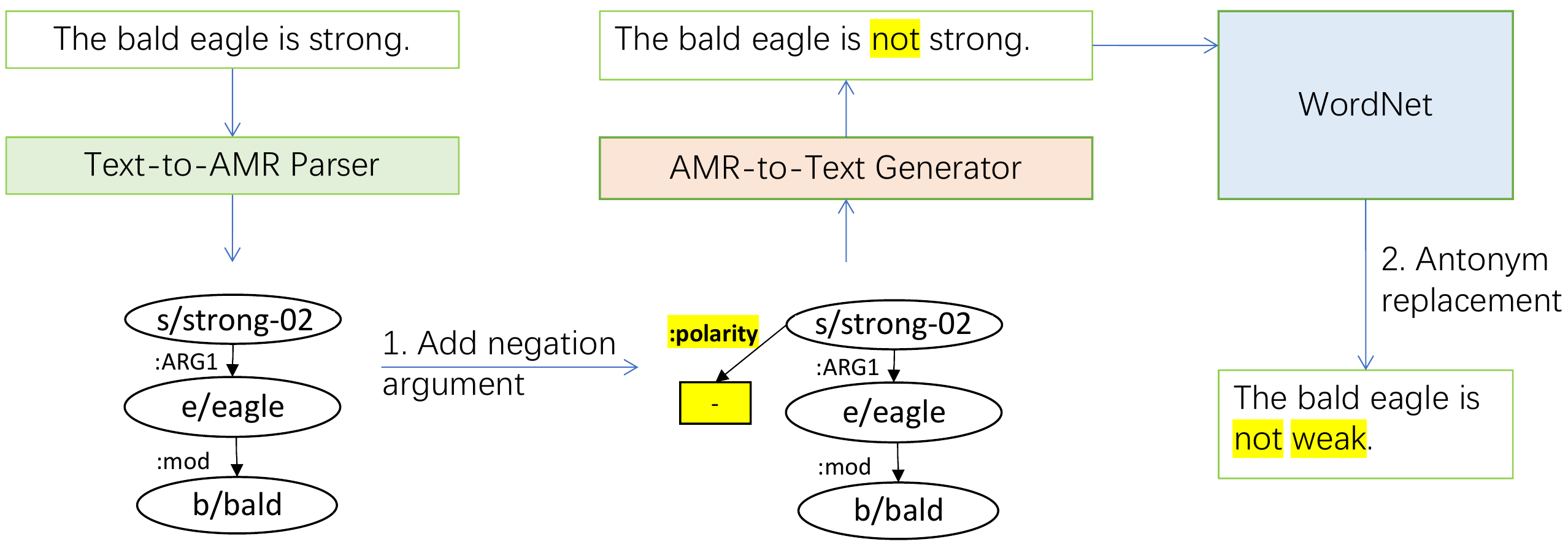}
\caption{An example for our AMR-based logic-driven data augmentation method using double negation law}
\label{double-negation-example}
\end{figure*}

\begin{figure*}
\begin{adjustbox}{max width=\textwidth}

\fbox{%
  \parbox{\textwidth}{%
  
\textit{Context (Facts+Rules):}

\textit{Facts:}
Alan is tall. Alan is big. Alan is huge. Fiona is thin. Fiona is small. Charlie is quiet. Charlie is smart. Charlie is wealthy. Anne is dull. Anne is sad. Anne is poor. 

\textit{Rules for Depth=1:}
If someone is \colorbox{green}{\color{black}tall} then they are \colorbox{green}{\color{black}quiet}. If someone is \colorbox{lime}{\color{black}thin} then they are \colorbox{lime}{\color{black}little}. If someone is \colorbox{cyan}{\color{black}dull and sad} then they are bad. If someone is quiet and smart then they are kind.

\textit{Rules for Depth=1 (with altered rules:}
If someone is \colorbox{green}{\color{black}not quiet} then they are \colorbox{green}{\color{black}not tall}. If someone is \colorbox{lime}{\color{black}not little} then they are \colorbox{lime}{\color{black}not thin}. If someone is \colorbox{cyan}{\color{black}sad and dull} then they are bad. If someone is smart and quiet then they are kind.

\textit{Question 1:} Alan is quiet? \textit{Label:} True.

\textit{Question 2:} Alan is not smart? \textit{Label:} False.

\textit{Question 3:} Fiona is little? \textit{Label:} True.

\textit{Question 4:} Fiona is not little? \textit{Label:} 
False.

\textit{Question 5:} Charlie is kind? \textit{Label:} True.

\textit{Question 6:} Charlie is not kind? \textit{Label:} False.

\textit{Question 7:} Anne is bad? \textit{Label:} True.

\textit{Question 8:} Anne is not bad? \textit{Label:} False.

  }%
}
\end{adjustbox}
\caption{An example for PARARULE-Plus Depth=1 and Depth=1 (with altered rules). The input includes context (facts + rules) and questions. The output is either true or false. In this example, we use logical equivalence laws (contraposition and commutative laws to extend the sentence in the rule sets to logical equivalence sentences. The highlighted words are the logical equivalence laws that we used. The green and lime green background mean the sentences are constructed by contraposition law, and the cyan background means the sentences are constructed by commutative law.)
}	
\label{PARARULE-Plus-depth-1}
\end{figure*}

\begin{figure*}
\begin{adjustbox}{max width=\textwidth}

\fbox{%
  \parbox{\textwidth}{%
  
\textit{Context (Facts+Rules):}

\textit{Facts:}
Erin is strong. Erin is tall. Erin is huge. Dave is thin. Dave is short. Fiona is kind. Fiona is wealthy. Fiona is quiet. Bob is sad. Bob is poor. Bob is bad.

\textit{Rules for Depth=2:}
\colorbox{green}{\color{black}Strong} people are \colorbox{green}{\color{black}kind}. If someone is thin and short then they are little. If someone is sad and poor then they are dull. If someone is \colorbox{lime}{\color{black}kind and wealthy} then they are \colorbox{lime}{\color{black}nice}. \colorbox{pink}{\color{black}All} little people are \colorbox{pink}{\color{black}small}. All kind people are wealthy. All nice people are smart. \colorbox{magenta}{\color{black}All} dull people are \colorbox{magenta}{\color{black}rough}.

\textit{Rules for Depth=2 (with altered rules):}
If someone is \colorbox{green}{\color{black}not kind} then they are \colorbox{green}{\color{black}not strong}. If someone is thin and short then they are little. If someone is sad and poor then they are dull. If someone is \colorbox{lime}{\color{black}not nice} then they are \colorbox{lime}{\color{black}not both kind and wealthy}. \colorbox{pink}{\color{black}There are no} little people who are \colorbox{pink}{\color{black}not small}. All kind people are wealthy. All nice people are smart. \colorbox{magenta}{\color{black}There are no} dull people who are \colorbox{magenta}{\color{black}not rough}.

\textit{Question 1:} Erin is wealthy? \textit{Label:} True.

\textit{Question 2:} Erin is not wealthy? \textit{Label:} False.

\textit{Question 3:} Dave is small? \textit{Label:} True.

\textit{Question 4:} Dave is not small? \textit{Label:} 
False.

\textit{Question 5:} Fiona is smart? \textit{Label:} True.

\textit{Question 6:} Fiona is not smart? \textit{Label:} False.

\textit{Question 7:} Bob is rough? \textit{Label:} True.

\textit{Question 8:} Bob is not rough? \textit{Label:} False.

  }%
}
\end{adjustbox}
\caption{An example for PARARULE-Plus Depth=2 and Depth=2 (with altered rules). The input includes context (facts + rules) and questions; the output is either ``True'' or ``False''. In this example, we use the contraposition law and De Morgan's law to convert sentences in the rule set to logically equivalent sentences. We highlighted the keywords that were changed when the alternative rules were constructed. Green and lime green backgrounds indicate sentences constructed using the contraposition law, while pink and magenta indicate sentences constructed with De Morgan's law.)
}
\label{PARARULE-Plus-depth-2}
\end{figure*}

Here are some synthetic sentence examples and more details for implication, conjunction, disjunction, and negation in the context of AMR-LDA mentioned in Algorithm~\ref{leamr-data-augmentation}.

\textbf{Double Negation Law}: The original sentence ``The bald eagle is strong'' is parsed into an AMR graph using a text-to-AMR parser. The parser confirms no negative meanings. To apply the double negation law, negative polarity is added, and an AMR-to-text generator then reforms the sentence. WordNet replaces the adjective with its antonym, creating a logically equivalent sentence.

\textbf{Commutative Law}: The sentence ``The bald eagle is clever and the wolf is fierce'' is converted into an AMR graph. The root node ``a/and'' of this graph, a conjunction argument, allows for the application of the commutative law by swapping arguments. The AMR-to-text generator then produces a new sentence, maintaining logical equivalence.

\textbf{Implication Law}: The sentence ``If Alan is kind, then Bob is clever'' is parsed into an AMR graph. The method checks for conditional and conclusion arguments. An ``or'' disjunction replaces the root node, and negative polarity is added to the first half of the sentence. The modified graph is then transformed back into a natural language sentence, ensuring logical equivalence.

\textbf{Contraposition Law}: The same initial sentence ``If Alan is kind, then Bob is clever'' is analyzed. The contraposition law is applied by swapping the conditional and conclusion arguments in the AMR graph and adding negative modifiers to both. The adjusted graph is then converted back into a logically equivalent sentence.

\begin{algorithm*}[h]
	\caption{AMR-Based Logic-Driven Data Augmentation} 
	\label{leamr-data-augmentation} 
	\begin{algorithmic}
		\REQUIRE Synthetic sentence lists (list1, list2, list3, and list4) generated following the patterns from \\Table \ref{logic-pattern-double-negation}, \ref{logic-pattern-commutative-law}, \ref{logic-pattern-contraposition-law}, and \ref{logic-pattern-implication-law} respectively.
		total\_list = []
		\FOR{sent in synthetic\_ sentence\_lists}{
        \item amr\_graph = Text-To-AMR-Parser(sent)
        \IF{sent in list1}{\item \emph{\textbf{\#\#double negation law}} \IF{``:polarity -'' in amr\_graph}{\item Remove ``:polarity -'' from the amr\_graph} \ELSE{}{\item Add ``:polarity -'' into the amr\_graph}\ENDIF}
        \item aug\_text = AMR-To-Text-Generator(amr\_graph)
        \item Use WordNet to replace an adjective word to antonym word from aug\_text.
		\ELSIF{sent in list2}{\item \emph{\textbf{\#\#commutative law}} \item Switch the order of two arguments.
        \item aug\_text = AMR-To-Text-Generator(amr\_graph)}
        \ELSIF{sent in list3}{\item \emph{\textbf{\#\#implication law}} \item Change the root node as ``or''. \IF{``:polarity -'' in a condition argument}{\item Remove the ``:polarity -''.} \ELSE{}{\item Add ``:polarity -'' into the argument.}\ENDIF
        \item aug\_text = AMR-To-Text-Generator(amr\_graph)}
        \ELSIF{sent in list4}{\item 
        \emph{\textbf{\#\#contraposition law}}  \item Switch the order of two arguments. \IF{``:polarity -'' in the argument of the amr\_graph}{\item Remove the ``:polarity -''.} \ELSE{}{\item Add ``:polarity -'' into the argument.}\ENDIF
        \item aug\_text = AMR-To-Text-Generator(amr\_graph)}\ENDIF \\
		\item total\_list = total\_list.append((sent, aug\_text, 1))}
		\ENDFOR
		\\
		return total\_list
	\end{algorithmic} 
\end{algorithm*}

\begin{algorithm*}[h]
	\caption{Negative samples construction} 
	\label{negative-sample-construction} 
	\begin{algorithmic}
		\REQUIRE Synthetic sentence lists (list1, list2, list3, and list4) generated following the patterns from \\Table \ref{logic-pattern-double-negation}, \ref{logic-pattern-commutative-law}, \ref{logic-pattern-contraposition-law}, and \ref{logic-pattern-implication-law} respectively.
		total\_list = [], total\_list2 = []
		\FOR{sent in synthetic\_ sentence\_lists}{
        \item amr\_graph = Text-To-AMR-Parser(sent)
        \IF{``:polarity -'' in amr\_graph}{\item Remove ``:polarity -''} \ELSE{}{\item Add ``:polarity -'' into the amr\_graph}\ENDIF
        \item aug\_text = AMR-To-Text-Generator(amr\_graph) \\
		\item total\_list = total\_list.append((sent, aug\_text, 0))}
        \FOR{sent in total\_list}{\item random select an index i from total\_list
        \item total\_list2 = total\_list2.append((sent, total\_list[i], 0))
        } \ENDFOR
		\ENDFOR
        \\
        total\_list = total\_list.extend(total\_list2)
        \\
		return total\_list
        \\
	\end{algorithmic} 
\end{algorithm*}

\begin{algorithm*}[h]
	\caption{Logical-Equivalence-Identification Contrastive Learning} 
	\label{representation-learning} 
	\begin{algorithmic}
		\REQUIRE positive\_list and negative\_list from Algorithm \ref{leamr-data-augmentation} and \ref{negative-sample-construction}, pre-trained large language model (LLM),\\ stage-2 downstream task datasets (ReClor, LogiQA, MNLI, RTE, QNLI, QQP), batch\_size bs, learning\_rate lr\\
		\emph{\textbf{Stage-1 fine-tuning}} \\
        \FOR{sents, pos\_sents, neg\_sents from zip(positive\_list, negative\_list, bs)}{
        \item LLM, Loss = Contrastive_loss(LLM, \\sents, pos\_sents, neg\_sents, label, lr)}\ENDFOR \\
        \emph{\textbf{Stage-2 fine-tuning}} \\
        \FOR{sent1, sent2 from zip(downstream\_tasks, bs)}{
        \item LLM, Loss = Cross\_entropy\_loss(LLM, sent1, sent2, label, lr)}\ENDFOR \\
	\end{algorithmic} 
\end{algorithm*}

\section{Confidence Intervals for the Main Experiments}\label{confidence-intervals}
Here are the confidence intervals for the main experiments in Table~\ref{Reclor-logiqa-confidence-intervals}. We select random seed 0, 21 and 42 to conduct the main experiment on ReClor and LogiQA datasets as shown on Table~\ref{Reclor-logiqa-confidence-intervals}. We utilise a 95\% confidence interval to calculate.

\begin{table*}[h]
\resizebox{1\textwidth}{!}{
\begin{tabular}{@{}lcccc@{}}
\toprule
Model/Datasets          & \multicolumn{4}{c}{ReClor}                                                                                                                            \\ \cmidrule(l){2-5} 
                        & Dev                                 & Test                                & Test-E                              & Test-H                              \\ \midrule
RoBERTa                 & 59.73 {[}54.83,64.00{]}          & 53.20 {[}52.30,54.00{]}          & 72.57 {[}69.50,75.00{]}          & 37.97 {[}34.30,41.00{]}          \\
RoBERTa LReasoner-LDA   & 59.46 {[}57.40,61.00{]}          & 53.66 {[}52.40,54.00{]}          & 72.19 {[}70.40,74.00{]}          & 39.10 {[}36.20,42.00{]}          \\
RoBERTa AMR-DA          & 58.66 {[}53.90,63.00{]}          & 53.93 {[}51.70,56.00{]}          & 66.81 {[}64.20,69.00{]}          & \textbf{43.80 {[}41.70,45.00{]}} \\
RoBERTa AMR-LDA         & \textbf{65.26 {[}60.50,70.00{]}} & \textbf{56.86 {[}55.20,58.00{]}} & \textbf{77.34 {[}73.90,80.00{]}} & 40.77 {[}39.80,41.00{]}          \\
DeBERTaV2               & 73.93 {[}66.20,81.00{]}          & 70.46 {[}60.80,80.00{]}          & 80.82 {[}76.50,85.00{]}          & 62.31 {[}47.70,77.00{]}          \\
DeBERTaV2 LReasoner-LDA & 75.73 {[}68.40,83.00{]}          & 70.70 {[}59.50,81.00{]}          & 84.08 {[}77.30,90.00{]}          & 60.17 {[}45.50,74.00{]}          \\
DeBERTaV2 AMR-DA        & 79.06 {[}73.60,84.00{]}          & 75.90 {[}73.70,78.00{]}          & 84.62 {[}80.20,89.00{]}          & 69.04 {[}66.20,71.00{]}          \\
DeBERTaV2 AMR-LDA       & \textbf{79.40 {[}77.60,81.00{]}} & \textbf{77.63 {[}73.80,81.00{]}} & \textbf{85.75 {[}83.20,88.00{]}} & \textbf{71.24 {[}66.40,76.00{]}} \\ \midrule
Model/Datasets          & \multicolumn{4}{c}{LogiQA}                                                                                                                            \\ \cmidrule(l){2-5} 
                        & \multicolumn{2}{c}{Dev}                                                   & \multicolumn{2}{c}{Test}                                                  \\ \midrule
RoBERTa                 & \multicolumn{2}{c}{35.43 {[}30.60,40.00{]}}                            & \multicolumn{2}{c}{34.50 {[}30.60,38.00{]}}                            \\
RoBERTa LReasoner-LDA   & \multicolumn{2}{c}{34.81 {[}31.60,39.00{]}}                            & \multicolumn{2}{c}{34.81 {[}30.90,38.00{]}}                            \\
RoBERTa AMR-DA          & \multicolumn{2}{c}{36.45 {[}29.40,44.00{]}}                            & \multicolumn{2}{c}{37.22 {[}34.50,41.00{]}}                            \\
RoBERTa AMR-LDA         & \multicolumn{2}{c}{\textbf{40.29 {[}36.40,47.00{]}}}                   & \multicolumn{2}{c}{\textbf{38.14 {[}34.50,41.00{]}}}                   \\
DeBERTaV2               & \multicolumn{2}{c}{39.72 {[}22.70,53.00{]}}                            & \multicolumn{2}{c}{39.62 {[}18.40,54.00{]}}                            \\
DeBERTaV2 LReasoner-LDA & \multicolumn{2}{c}{30.87 {[}30.30,31.00{]}}                            & \multicolumn{2}{c}{28.51 {[}21.80,36.00{]}}                            \\
DeBERTaV2 AMR-DA        & \multicolumn{2}{c}{29.95 {[}25.40,36.00{]}}                            & \multicolumn{2}{c}{30.10 {[}27.30,32.00{]}}                            \\
DeBERTaV2 AMR-LDA       & \multicolumn{2}{c}{\textbf{42.34 {[}36.70,48.00{]}}}                   & \multicolumn{2}{c}{\textbf{39.88 {[}25.70,49.00{]}}}                   \\ \bottomrule
\end{tabular}}
\caption{The confidence intervals for the main experiments conducted on the ReClor and LogiQA datasets. We select random seed 0, 21 and 42 to conduct the main experiment on ReClor and LogiQA datasets. We utilise a 95\% confidence interval to calculate the confidence interval.}
\label{Reclor-logiqa-confidence-intervals}
\end{table*}

\end{document}